\documentclass[10pt,twocolumn,letterpaper,dvipsnames,table]{article}

\usepackage{cvpr}              
\usepackage{times}
\usepackage{svg}
\usepackage{multirow} 
\usepackage{makecell}
\usepackage{CJKutf8}         
\usepackage{rotating}  
\usepackage{graphicx}
\usepackage{marvosym} 
\usepackage{footmisc} 
\usepackage{fontawesome} 
\usepackage{ragged2e}  

\newif\ifshowcomments
\showcommentstrue

\ifshowcomments

\definecolor{cvprblue}{rgb}{0.21,0.49,0.74}
\usepackage[pagebackref,breaklinks,colorlinks,allcolors=cvprblue]{hyperref}

\def\paperID{10237} 
\def\confName{ICCV}
\def\confYear{2025}

\title{GMAI-VL \& GMAI-VL-5.5M: A Large Vision-Language Model and \\ A Comprehensive Multimodal Dataset Towards General Medical AI}

\author{
Tianbin Li$^{1}$\footnotemark[1]\ , Yanzhou Su$^{1}$\thanks{Equal contribution}\ , Wei Li\textsuperscript{1,2}, Bin Fu$^{1,3}$, Zhe Chen$^{1,4}$, Ziyan Huang$^{1,2}$ \\
Guoan Wang$^{1,5}$, Chenglong Ma$^{1,6}$, Ying Chen$^{1,7}$, Ming Hu$^{1,8}$, Yanjun Li$^{1,5}$, Pengcheng Chen$^{1,9}$,\\
Xiaowei Hu$^{1}$, Zhongying Deng$^{1,10}$, Yuanfeng Ji$^{11}$, Jin Ye$^{1,8}$, Yu Qiao$^{1}$, Junjun He$^{1}$\thanks{Corresponding author (hejunjun@pjlab.org.cn)} \\
$^{1}$Shanghai Artificial Intelligence Laboratory \quad $^{2}$Shanghai Jiao Tong University \\ 
\quad $^{3}$Shenzhen Institute of Advanced Technology (SIAT), Chinese Academy of Sciences \\
$^{4}$Nanjing University \quad $^{5}$East China Normal University  \quad $^{6}$Fudan University \\
$^{7}$Xiamen University   \quad $^{8}$Monash University \quad $^{9}$University of Washington \\ 
\quad $^{10}$University of Cambridge  \quad $^{11}$Stanford University \\ 
}

\begin{document}
\maketitle


\begin{abstract}

Despite significant advancements in general AI, its effectiveness in the medical domain is limited by the lack of specialized medical knowledge. 
To address this, we formulate GMAI-VL-5.5M, a multimodal medical dataset created by converting hundreds of specialized medical datasets with various annotations into high-quality image-text pairs. 
This dataset offers comprehensive task coverage, diverse modalities, and rich image-text data. 
Building upon this dataset, we develop GMAI-VL, a general medical vision-language model, with a three-stage training strategy that enhances the integration of visual and textual information. 
This approach significantly improves the model's ability to process multimodal data, supporting accurate diagnoses and clinical decision-making. 
Experiments show that GMAI-VL achieves state-of-the-art performance across various multimodal medical tasks, including visual question answering and medical image diagnosis.

\end{abstract}    
\section{Introduction}

Recent advancements in Large-scale Vision-Language Models (LVLMs) have driven progress in image recognition, natural language processing, and multimodal tasks, leveraging the power of multimodal datasets. 
In the medical field (general medical AI, GMAI), as these technologies mature, the need for accurate processing of diverse data—such as medical images, clinical text, and structured records—has become critical for reliable diagnostic and treatment decisions.

However, existing LVLMs, like GPT-4o~\citep{achiam2023gpt}, face limitations in medical applications due to their lack of domain-specific knowledge. 
This highlights the need for specialized solutions that incorporate medical expertise. To address this, we develop a comprehensive medical vision-language dataset and corresponding domain-specific models.

\begin{figure*}[tp]
    \centering
    \includegraphics[width=\textwidth]{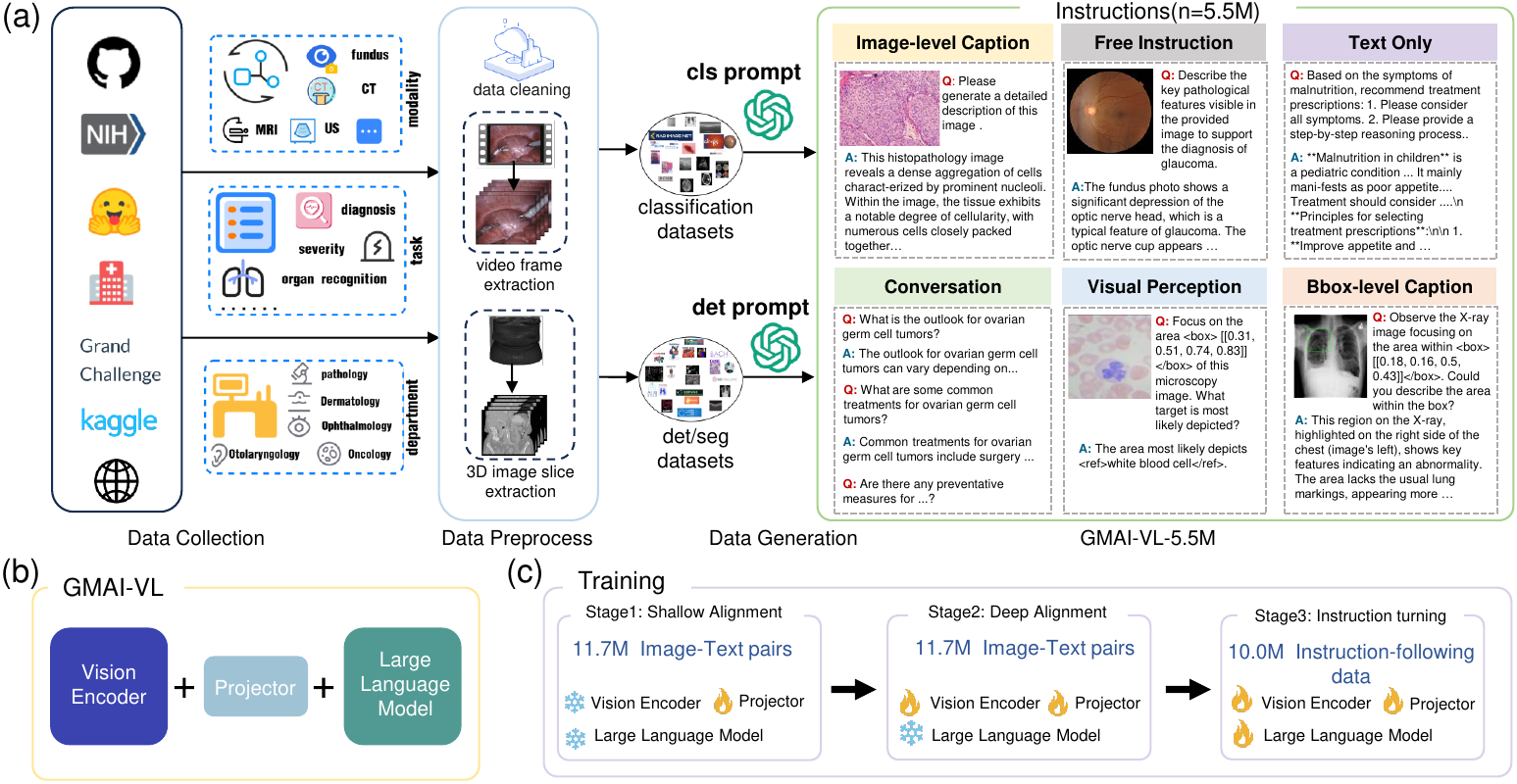}
    \caption{Overview of GMAI-VL and GMAI-VL-5.5M. (a) Sources, departments, modalities, task types, and instruction formats of GMAI-VL-5.5M. (b) Architecture of GMAI-VL, with a Vision Encoder, Projector, and Large Language Model. (c) Three-stage training process, including shallow alignment, deep alignment, and instruction tuning, with corresponding data sizes and components.  \includegraphics[width=0.018\textwidth]{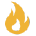} indicates the training part while \includegraphics[width=0.018\textwidth]{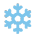} indicates the frozen part.}
    \label{fig:pipeline}
\end{figure*}

As shown in Fig.~\ref{fig:pipeline} (a), Our dataset provides high-quality medical knowledge across three aspects:
(i) \textit{Comprehensive medical tasks}: 
To improve the model's applicability in various medical scenarios, our dataset covers a wide range of disease types, symptoms, treatments, and medical workflows. 
(ii) \textit{Rich multimodal representation}: Our dataset includes a variety of modalities, such as different types of medical images (e.g., CT, MRI, X-rays) and textual data (e.g., medical records, imaging reports). 
(iii) \textit{High-quality image-text data}: The performance of medical LVLMs heavily relies on high-quality image-text pairs. We use the well-curated collection of medical images paired with precise textual descriptions to build our dataset.

To achieve this, we develop an annotation-guided data generation process to create this large-scale multimodal medical dataset.
First, we collect multiple open-source medical imaging datasets and extract their key annotations (\textit{e.g.}, modality, task type, labels, bounding boxes).
Next, we convert these well-labeled annotations into text descriptions suitable for training vision-language models.
This ensures that the medical image-text pairs are accurately constructed, with annotations verified by medical experts in the collected medical imaging datasets.
The process results in the GMAI-VL-5.5M dataset, consisting of 5.5 million samples, which supports the development of general medical LVLMs.

Building upon this dataset, we develop a general medical vision-language model, GMAI-VL. 
To optimize its ability to integrate visual and linguistic features and follow complex instructions, we present a three-stage training strategy. The first two stages involve shallow and deep alignments, gradually establishing connections between medical images and texts, from basic features to high-level semantics. In the final stage, the model is fine-tuned with cross-modal instructions, improving its understanding of visual-language interactions and its ability to follow intricate tasks, as shown in Fig.~\ref{fig:pipeline} (b)\&(c).

We conduct experiments on various multimodal medical benchmarks~\citep{lau2018dataset,he2020pathvqa,liu2021slake,zhang2023pmc,chen2024gmai,hu2024omnimedvqa,yue2024mmmu}, and the results demonstrate that GMAI-VL significantly outperforms existing models. 
Specifically, GMAI-VL sets new benchmarks with an average score of 88.48\% on OmniMedVQA, 62.43\% on the GMAI-MMBench \textit{test} set, and 51.3\% on the Health \& Medicine track of MMMU.

The contributions of this work are as follows:

\begin{itemize}[leftmargin=*]

\item We develop an annotation-guided data generation methodology to create GMAI-VL-5.5M, a comprehensive medical vision-language dataset with diverse medical tasks, rich multimodal representations, and high-quality image-text pairs.

\item Leveraging this dataset, we design GMAI-VL, a general medical vision-language model, and propose a three-stage training strategy to enhance its integration of visual and linguistic features, improving its performance in various medical tasks.

\item We demonstrate that GMAI-VL outperforms existing models in multimodal question-answering tasks, setting new benchmarks with high performance on OmniMedVQA, GMAI-MMBench, and the Health \& Medicine track of MMMU.

\end{itemize}

\section{Related Work}

\textbf{Large-scale medical vision-language datasets.} 
These datasets are crucial for training Large Vision-Language Models (LVLMs) in the medical domain. While general datasets are readily available, biomedical datasets often focus on text or images separately, limiting their generalization. 
Datasets like MIMIC-CXR \citep{johnson2019mimic} and CheXpert \citep{chambon2024chexpert} advance radiology models but are restricted to a single image modality (X-ray), limiting their use as general-purpose medical LVLMs.

To address this, researchers have compiled large-scale vision-language datasets by scraping public resources like PubMed and medical textbooks. 
Datasets like LLaVA-Med \citep{li2024llava}, Med-Flamingo \citep{moor2023med}, and PubMedVision \citep{chen2024huatuogpt} improve upon LLaVA-Med by enhancing the quality of medical data. 
Additionally, open-source image datasets with annotations have been converted into image-text pairs for training. 
Notable examples include RadFM \citep{wu2023towards}, MedDr \citep{he2024meddr}, BiomedGPT \citep{zhang2024generalist}, Med-Gemini \citep{saab2024capabilities}, and Med-PaLM \citep{singhal2023large}.
MedTrinity-25M~\citep{xie2024medtrinity} generates image-text pairs for supervised fine-tuning.

However, these datasets often face limitations in modalities, data sources, or task coverage, necessitating improvements. We construct a comprehensive medical vision-language dataset with broad task coverage, diverse modalities, and high-quality image-text pairs to provide a solid foundation for model training.

\begin{table*}[tp]
\centering
\caption{Comparison of multimodal medical  datasets, including size, modality, language, traceability, and sources.}
\label{tab:comparison-LVLMs}
\resizebox{0.8\textwidth}{!}{
\begin{tabular}{l|ccccc}
\toprule
Datasets      & Data Size  & Modality  & Language & Traceability & Data Source       \\
\midrule
PathVQA \citep{he2020pathvqa} & 32.7k & Pathology  & EN & \texttimes  & Textbooks \\
MIMIC-CXR \citep{johnson2019mimic} & 227k & X-Ray  & EN & \checkmark & Hospital    \\ 
quilt-1M \citep{ikezogwo2024quilt}  &  1M & Pathology  & EN & \texttimes & YouTube \& PubMed\\
MedDr VQA \citep{he2024meddr}  &  197k  & Multimodal  & EN  & \checkmark & 13 medical datasets\\ 
PMC-OA \citep{lin2023pmc} & 1.65M & Multimodal  & EN & 
 \texttimes & PubMed \\
PMC-VQA \citep{zhang2023pmc} & 413k  &  Multimodal & EN  &  \texttimes & PubMed \\ 
LLaVA-Med VQA \citep{li2024llava}   & 56,702  & Multimodal  & EN & \texttimes &  PubMed  \\ 
ChiMed-VL \citep{liu2023qilin} & 1.05M  & Multimodal &   CN  & \texttimes  & PubMed  \\ 
PMC-CaseReport \citep{wu2023towards} & 438k & Multimodal & EN & \texttimes & PubMed   \\ PubMedVision \citep{chen2024huatuogpt} & 1.29M  & Multimodal  &  EN\&CN & \texttimes & PubMed   \\ \midrule 
\multirow{2}{*}{\textbf{GMAI-VL-5.5M (ours)}} & \multirow{2}{*}{5.5M}  & \multirow{2}{*}{Multimodal} &\multirow{2}{*}{EN\&CN} &   \multirow{2}{*}{ \checkmark} & \multirow{2}{*}{\makecell{219 specialized medical\\ imaging datasets}}   \\ 
\\
\bottomrule
\end{tabular} }%
\vspace{-2mm}
\end{table*}

\vspace{-2.5mm}
\paragraph{Medical vision-language models.} 
Medical vision-language models often adapt general-purpose Large Vision-Language Models (LVLMs) for specific medical tasks using specialized datasets. 
For example, Med-Flamingo~\citep{moor2023med} improves OpenFlamingo-9B with 0.8 million interleaved and 1.6 million paired medical image-text data for medical image analysis and report generation. 
RadFM~\citep{wu2023towards} enhances PMC-LLaMA~\citep{wu2023pmc} with 16 million radiology images and text descriptions. 
Med-PaLM~\citep{tu2024towards} adapts PaLM-E~\citep{driess2023palm} to medical data, achieving state-of-the-art results in diagnostic support and Q\&A. 
LLaVA-Med~\citep{li2024llava} uses a biomedical figure-caption dataset from PubMed Central to improve LLaVA~\citep{touvron2023llama1,touvron2023llama} for biomedical image understanding and open-ended conversations. 
Med-Gemini~\citep{saab2024capabilities} leverages long-format question-answering datasets for better multimodal and contextual capabilities, enhancing complex medical Q\&A and reasoning tasks. 
Additionally, HuatuoGPT-Vision~\citep{chen2024huatuogpt} and MedDr~\citep{he2024meddr} adapt general-purpose LVLMs like LLaVA and InternVL to various medical modalities, including radiology, pathology, dermatology, and endoscopy.

While these studies focus on constructing medical datasets, they often overlook adaptation strategies. 
Naive training methods may fail to bridge the gap between natural and medical image-text pairs, or align diverse medical modalities and texts (e.g., prescriptions, radiology reports, EHRs), limiting generalizability. Our work introduces a novel three-stage training strategy to better integrate visual and language features, improving generalization.

\begin{table*}[tp]
    \centering
    \caption{The examples of annotation-guided prompts for paired data format generation.}
    \label{tab:prompts}
    \begin{tabular}{p{\textwidth}}
        \toprule
        \multicolumn{1}{c}{\textbf{Instruction-following data prompt}} \\
        \midrule
        \begin{minipage}{1.0\textwidth} 
            \begin{minipage}{0.2\textwidth} 
                \centering
                \includegraphics[width=0.95\linewidth]{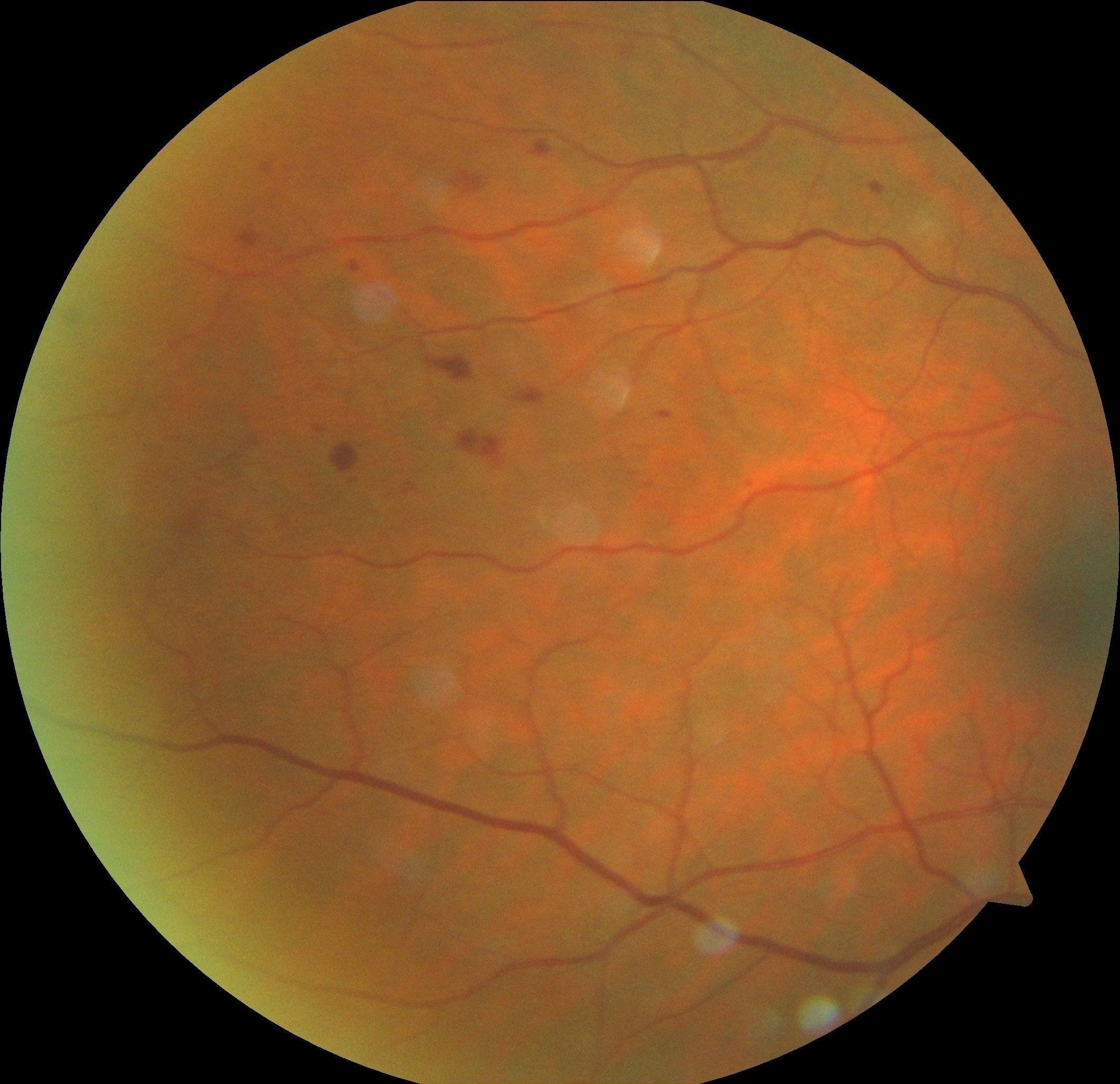}  
            \end{minipage} 
            \hfill
            \begin{minipage}{0.8\textwidth} 
                As a medical expert, you will receive a \textless fundus\textgreater image belonging to \textless Ophthalmology\textgreater, which is labeled as \textless moderate nonproliferative diabetic retinopathy\textgreater. \\
                Please generate question-answer pairs based on the following requirements:\\
                1. The question-answer pairs must be strictly based on the content of the image and directly related to the label information, should not state known medical facts or general definitions directly. \\
                2. Focus on reasoning around the label and guide observation of the visible features of the disease, such as affected areas, symptoms, and changes in the retina. \\
                3. Each question and answer must be precise, clear, and derived from observable characteristics in the image, with no vague or uncertain information. \\
                4. The output format must follow the standard JSON format with the structure as follows: \\
                \{ 'question': \textless question content \textgreater , 'answer': \textless answer content \textgreater \}
            \end{minipage}
        \end{minipage} \\
        \midrule
        \textbf{Question} \textit{Can we see any hard exudates or fluid leakage in the image?} \\
        \textbf{Answer} \textit{The image does show some bright, yellowish deposits, which are consistent with hard exudates, often seen in moderate nonproliferative diabetic retinopathy due to fluid leakage.}\\
        \textbf{Question} \textit{What is the appearance of the optic disc in this retinal image?} \\
        \textbf{Answer} \textit{The optic disc appears normal without any visible signs of swelling or abnormal changes, suggesting that the optic nerve is not yet affected by the disease.}\\
        \midrule
        \midrule 
        \multicolumn{1}{c}{\textbf{Caption data prompt}} \\
        \midrule
        \begin{minipage}{1.0\textwidth} 
            \begin{minipage}{0.2\textwidth} 
                \centering
                \includegraphics[width=\linewidth]{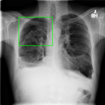}  
            \end{minipage} 
            \hfill
            \begin{minipage}{0.8\textwidth} 
                As a medical expert, you are given a \textless X-ray\textgreater\ image belonging to \textless Pulmonary Medicine\textgreater, where the \textless bbox [179,164,500,434]\textgreater\ is labeled as \textless Pneumothorax\textgreater. \\
                Please generate a very detailed description related to bbox in the given image. When generating the description, you must strictly adhere to the following guidelines: \\
                1. The generated description should be inferable from the image; \\
                2. The generated description should include image features related to the given label; \\
                3. The generated description should start with 'This region'; \\
                4. The description may imply known information but should not directly state known information; \\
                5. The generated description should include major entities within and around the region, background elements, texture features, and any other notable features. \\
            \end{minipage}
        \end{minipage} \\
        \midrule
        \textbf{Output:} This region shows ... \textit{the right hemithorax on a frontal chest X-ray}, ... \textit{an area suggestive of abnormal air accumulation}. \textit{The bounded area appears less dense compared to surrounding lung tissue}, ... \textit{The edges of the lung are visible, separated from the chest wall by a faint radiolucent (darker) space, suggestive of pneumothorax}. \textit{The lung markings are notably absent within this area}, ... \textit{The texture within the defined area has a homogeneous, darkened appearance consistent with air rather than tissue density, contrasting with the granular pattern in the aerated lung parenchyma outside this region.} \\
        \bottomrule
    \end{tabular}
\end{table*}

\section{GMAI-VL-5.5M: A Comprehensive Multimodal Medical Dataset}

With rapid advancements in medical vision-language models, constructing high-quality datasets is essential for developing general-purpose models. 
Unlike previous approaches that mainly rely on published literature, our method leverages specialized medical datasets with various annotations to create a more robust, high-quality resource.

We present GMAI-VL-5.5M, a comprehensive medical vision-language dataset that aggregates data from diverse open-source and proprietary sources. 
Covering 13 medical imaging modalities and 18 specialties, it supports a wide range of medical imaging tasks. 
This dataset enhances the model's ability to process complex medical information, advancing precision medicine and intelligent diagnostics.

\subsection{Data Curation}

\paragraph{Data collection.} 
To construct a comprehensive multimodal medical dataset, we sourced 219 datasets from diverse platforms. 
Fig.~\ref{fig:pipeline}(a) highlights key data sources, such as Kaggle, Grand Challenge, and Huggingface, which facilitate extensive data collection. 
These datasets cover diverse imaging modalities, including fundus, CT, MRI, and ultrasound (US), and span a variety of medical tasks, such as diagnosis, severity assessment, and organ recognition. 
They also encompass multiple clinical specialties, including pathology, dermatology, ophthalmology, otolaryngology, and oncology, further enhancing their diversity.

\vspace{-2.5mm}
\paragraph{Data processing.}
After data collection, we follow a workflow to pair medical data (including both 2D and 3D data) with corresponding text annotations.
To ensure high-quality annotations, we first extract key information from the annotations provided by medical experts. For classification data, we extract the modality, department, and labels for each image, discarding instances with missing or unclear annotations. 
For segmentation data, we follow the SA-Med2D-20M~\citep{ye2023sa} approach, filtering out low-quality images and labels, and converting them into detection annotation format.
Finally, the preprocessed data is standardized and organized into a structured format: \texttt{<image, modality, label, department, bbox [optional]>}, where ``bbox'' refers to the bounding box locations for detection annotations.

\vspace{-2.5mm}
\paragraph{Paired data format generation.}

To generate paired visual medical data with text descriptions, we use large vision-language models (\textit{i.e.}, GPT-4o) to produce detailed descriptions with instructions via an annotation-guided methodology. 
In details, for classification datasets, comprehensive descriptions of the entire image are created, while for detection datasets, the focus is on specific regions enclosed by bounding boxes, with detailed functional analyses of these areas. 
Furthermore, based on the given information, instruction-following question-answer pairs related to the medical images are generated. These pairs involve specific instructions tailored to the medical context, such as identifying critical anatomical or pathological features within the medical images (e.g., tumors, lesions, or organs) or providing in-depth interpretations of regions of interest. These instructions guide the model to produce relevant, precise responses, thereby enhancing its applicability to specialized medical imaging tasks.
As mentioned in the previous subsection, the segmentation dataset is transformed into a detection format using external bounding boxes, and data generation follows the detection dataset protocols.
The detailed example prompts are shown in Table~\ref{tab:prompts}.
The generated data is then used for medical Visual Question Answering (VQA) tasks, forming the comprehensive VQA dataset, GMAI-VL-5.5M. 
To enhance the model's multilingual capability, we translate approximately 30\% of the English image-text data into Chinese.
This multilingual data helps further improve the generalization ability of domain-specific multimodal models.

%

\subsection{Data Property}

\paragraph{Data statistics.}
Our dataset encompasses a broad spectrum of medical imaging tasks and modalities, forming a robust foundation for the development and evaluation of medical LVLMs.
Table~\ref{tab:distribution} summarizes the distribution of key modalities, tasks, clinical departments, and clinical tasks within GMAI-VL-5.5M, illustrating its diversity and extensive coverage. 
\textit{For a more detailed analysis, additional visual insights into the dataset composition can be found in the supplementary material.}
%

\vspace{-2.5mm}
\paragraph{Data quality.}
The quality of the generated paired data and annotations is ensured through two main approaches.
First, we use high-quality datasets from trusted sources, including professional challenges like Kaggle and GrandChallenge, as well as peer-reviewed datasets, ensuring data accuracy and reliability.
Second, we carefully control the data generation process. While GPT is used for generation, the prompts are designed with essential annotations (e.g., \texttt{<image, modality, label, department, bbox [optional]>}) to minimize errors. 
This annotation-guided approach produces more detailed and professional descriptions.
\textit{The data quality is detailed in the supplementary material.}


\begin{table}[tp]
\centering
\caption{Distribution of GMAI-VL-5.5M across key dimensions.}
\resizebox{0.98\linewidth}{!}{
\begin{tabular}{l|l|c}
\hline
\textbf{Dimension} & \textbf{Value} & \textbf{Percentage} \\
\hline
\multirow{3}{*}{\textbf{Task Type}} & 2D Classification & 50.4\% \\
& 3D Segmentation & 30.3\% \\
& 2D Segmentation & 12.7\% \\
& 2D Detection & 6.6\% \\
\hline
\multirow{5}{*}{\textbf{Modality}} & CT & 26.8\% \\
& MR & 24.7\% \\
& Endoscopy & 12.6\% \\
& Pathology & 11.2\% \\
& X-Ray & 6.7\% \\
& Fundus & 5.3\% \\
& Ultrasound & 3.0\% \\
\hline
\multirow{5}{*}{\textbf{Department}} & Orthopedic Surgery & 12.9\% \\
& General Surgery & 10.3\% \\
& Gastroenterology & 9.7\% \\
& Hematology & 9.2\% \\
&  Pulmonary Medicine & 9.0\% \\
&  Sports Medicine & 8.2\% \\
\hline
\multirow{6}{*}{\textbf{Clinical Task}} & Disease Diagnosis & 40.4\% \\
& Organ Recognition & 16.0\% \\
& Bone Recognition & 8.5\% \\
& Severity Grading & 6.1\% \\
& Surgeon Action Recognition & 6.0\% \\
\hline
\end{tabular}}
\label{tab:distribution}
\vspace{-5mm}
\end{table}

\vspace{-2.5mm}
\paragraph{Compared with other medical multimodal datasets.} 
The GMAI-VL-5.5M dataset, as shown in Table~\ref{tab:comparison-LVLMs}, distinguishes itself with its unmatched scale, featuring over 5.5 million samples from more than 219 specialized medical imaging datasets. 
Unlike other datasets, GMAI-VL-5.5M supports a broader range of modalities and languages, making it a truly global resource that addresses diverse clinical needs. 
Furthermore, GMAI-VL-5.5M ensures data traceability, maintaining high clinical relevance and reliability. 
This extensive and varied dataset is crucial for advancing medical multimodal research, enabling more effective training of LVLMs that can generalize across numerous medical tasks and scenarios, ultimately driving innovations in precision medicine and intelligent diagnostics.

\section{GMAI-VL: A General Medical Vision-Language Model}

\subsection{Architecture}

The GMAI-VL model is a vision-language model built upon the LLaVA architecture~\citep{liu2023llava,li2024llava}, incorporating three key components: a large language model (LLM), a vision encoder, and a projector (MLP), as illustrated in Fig.~\ref{fig:pipeline}(b). 


We use InternLM2.5-7B~\citep{team2023internlm} as our language processing module, which provides exceptional reasoning capabilities. With a context window of up to one million tokens, it can manage complex medical tasks and generate coherent, accurate responses. 
For vision processing, we adopt a CLIP-based vision encoder~\citep{radford2021learning}, which converts visual inputs into high-dimensional feature representations. 
The MLP projector bridges the vision encoder and language processing module, optimizing high-dimensional outputs and improving feature representation. 
This framework enables GMAI-VL to effectively process multimodal medical data.

\subsection{Training Strategy}
\label{sec:subtraining}

As shown in Fig.\ref{fig:pipeline}(c), the training process of the GMAI-VL model is divided into three stages: shallow alignment, deep alignment, and instruction tuning. 
To enhance the training of GMAI-VL, we supplement the GMAI-VL-5.5M dataset with additional medical datasets, resulting in a total training dataset of 11.7M samples.  

\vspace{-2.5mm}
\paragraph{Stage I: Shadow alignment.}
In this phase, we use a large-scale medical image-text dataset of approximately 11.7 million image-text pairs.
During this stage, we freeze both the large language model and the vision encoder, optimizing only the projector. 
This optimization establishes an initial alignment between medical images and their corresponding textual descriptions. 

During training, (i) the objective is to minimize the cross-entropy loss of the text tokens; (ii) all images are resized to $336 \times 336$ pixels; (iii) the learning rate is set to \(1e^{-3}\) with a cosine decay schedule, and AdamW is used as the optimizer; (iv) the total batch size is \(32 \times 8 \times 2\), where 32 refers to the number of GPUs used, eight represents the micro-batch size per GPU, and two is the number of gradient accumulation steps; and (v) a soft packing technique is used to allow each sample to contain multiple sequences, averaging over two sequences per sample.

\vspace{-2.5mm}
\paragraph{Stage II: Deep alignment.}
Most vision encoders in multimodal models are pre-trained on natural images, so addressing the domain gap between medical and natural images is critical during the deep alignment stage. 
To bridge this gap, we fine-tune both the vision-language projector and the vision encoder, enhancing the alignment between the visual features of medical images and the language model's feature space. 
The learning rate is set to \(1e^{-4}\), the total batch size is \(32 \times 4 \times 4\), and other settings remain consistent with Stage I.

\vspace{-2.5mm}
\paragraph{Stage III: Instruction tuning.}
In this stage, we fine-tune the entire GMAI-VL model—vision encoder, language model, and projector—through instruction tuning to enhance its instruction-following and dialogue capabilities. 
The multimodal instruction data is primarily derived from the previous stages. 
Note that this stage discards the low-quality samples that with too short descriptions or extremely lower confidence in the previous stages.
Additionally, medical text dialogue data is incorporated to improve the model's handling of various dialogue scenarios. This results in ten million samples for instruction tuning.
During taining, the learning rate is set to \(1e^{-5}\), the total batch size is \(32 \times 4 \times 4\), while other parameters, including optimizer, remain unchanged across stages. 

%

%

\begin{table}[tp]
\centering
\caption{Results on medical VQA benchmarks. The highest performance in each column is highlighted in \textcolor{red}{red}, and the second-highest performance is highlighted in \textcolor{blue}{blue}.}
\resizebox{\linewidth}{!}{
\begin{tabular}{lccccc}
\toprule
\textbf{Model} & \textbf{VQA-RAD} & \textbf{SLAKE}  & \textbf{PMC-VQA} & \textbf{Avg.} \\
\midrule
Med-Flamingo \citep{moor2023med} & 45.4 & 43.5 &  23.3 & 37.4 \\
RadFM \citep{wu2023towards} & 50.6 & 34.6 & 25.9 & 37.0 \\
LLAVA-Med-7B \citep{li2024llava} & 51.4 & 48.6 & 24.7 & 41.6 \\
Qwen-VL-Chat \citep{bai2023qwen} & 47.0 & 56.0 & 36.6 & 46.5 \\
Yi-VL-34B \citep{young2024yi} & 53.0 & 58.9  & 39.5 & 50.5 \\
LLAVA-v1.6-7B \citep{liu2024llavanext} & 52.6 & 57.9 & 35.5 & 48.7 \\
LLAVA-v1.6-13B \citep{liu2024llavanext} & 55.8 & 58.9  & 36.6 & 50.8 \\
LLAVA-v1.6-34B \citep{liu2024llavanext} & 58.6 & 67.3 & 44.4 & 56.8 \\
HuatuoGPT-Vision-7B \citep{chen2024huatuogpt} & \textcolor{blue}{63.8} & \textcolor{red}{74.5}  & \textcolor{blue}{52.7} & \textcolor{blue}{\textbf{63.7}} \\ \midrule
\textbf{GMAI-VL (w/o our data)} & 62.3 & 66.3  & 39.0 &  55.9 \\
\textbf{GMAI-VL (ours)} & \textcolor{red}{66.3} & \textcolor{blue}{72.9}  & \textcolor{red}{54.3} &  \textcolor{red}{64.5}\\
\bottomrule
\end{tabular} }
\label{tab:tradi_result}%
\vspace{-3mm}
\end{table}

\section{Experimental Results}

To evaluate our model, we utilized several established multimodal medical benchmarks, including medical VQA benchmarks (PMCVQA~\citep{zhang2023pmc}, PathVQA~\citep{he2020pathvqa}, VQA-RAD~\citep{lau2018dataset}, and SLAKE~\citep{liu2021slake}) as well as benchmarks designed for large vision-language models (OmniMedVQA~\citep{hu2024omnimedvqa}, GMAI-MMBench~\citep{chen2024gmai}, and the MMMU Health \& Medicine track~\citep{yue2024mmmu}). 
These benchmarks target specific aspects of medical image understanding and question answering. 
\textit{Detailed information about these benchmarks can be found in the supplementary material.}

We evaluate model performance using VLMEvalKit~\cite{duan2024vlmevalkit} with its default settings. To prevent data leakage in GMAI-VL-5.5M, we ensure that the evaluation datasets, OmniMedVQA and GMAI-MMBench, are sourced exclusively from public test sets, while the training data is strictly from public training sets. Additionally, MD5 hashes were computed for each image to verify that there are no duplicates with the benchmark images.

\subsection{Comparisons on Medical VQA Datasets}

The performance of various VLMs on popular medical VQA benchmark datasets is summarized in Table~\ref{tab:tradi_result}. 
Our model, GMAI-VL (ours), demonstrates strong performance, achieving the highest score of 66.3\% on the VQA-RAD~\citep{lau2018dataset} dataset, outperforming models such as HuatuoGPT-Vision-7B. This result underscores GMAI-VL's superior capability in handling radiological image question-answering tasks. On the PMC-VQA~\citep{zhang2023pmc} dataset, GMAI-VL achieves 54.3\%, and 72.9\% on SLAKE~\citep{liu2021slake}, highlighting its effectiveness across diverse medical VQA tasks. 
Overall, GMAI-VL demonstrates competitive performance across multiple benchmarks, showcasing its versatility in medical image understanding and question answering.

\subsection{Comparisons on OmniMedVQA}

\begin{table}[tp]
  \centering
  \caption{Comparison of LVLMs and GMAI-VL on OmniMedVQA across five question types. 
  The best performance in each column is highlighted in \textcolor{red}{red}, and the second-best in \textcolor{blue}{blue}. \textbf{Abbreviations}: MR = Modality Recognition, AI = Anatomy Identification, DD = Disease Diagnosis, LG = Lesion Grading, OBA = Other Biological Attributes.}
  \setlength{\tabcolsep}{3.5pt} 
  \resizebox{\linewidth}{!}{
    \begin{tabular}{l|c|c|c|c|c|c}
    \toprule[1pt]
    \textbf{Model} & \makecell{\textbf{MR}}  & \makecell{\textbf{AI}} & \makecell{\textbf{DD}} & \makecell{\textbf{LG}} & \makecell{\textbf{OBA}} & \makecell{\textbf{Overall}} \\
    \hline
    Random Guess & 25.00 & 25.84 & 28.41 & 25.40 & 37.49 & 28.28\\
    \multicolumn{7}{c}{\cellcolor[HTML]{FFF3E4}Open-Source LVLMs}\\  
    MiniGPT-4 \citep{zhu2023minigpt}   &  36.98 & 32.68 & 24.19 & 20.45 & 26.14 & 27.59 \\
    LLaVA \citep{liu2023llava} &  52.30 & 35.27 & 11.80 & 9.77 & 24.70 & 22.86 \\
    LLaMA\_Adapter\_v2 \citep{gao2023llama}   &   58.45 & 38.18 & 29.12 & 23.73 & 30.97 & 35.08 \\
    InstructBLIP \citep{dai2024instructblip}   &  72.35 & 39.90 & 32.01 & 43.80 & 47.91 & 41.14   \\
    BLIP-2 \citep{li2023blip}  &  57.48 & 49.83 & 46.21 & 30.52 & 73.52 & 50.77 \\ 					
    Qwen-VL-Chat \citep{bai2023qwen} & 33.69 & 10.95 & 16.27 & 6.71 & 41.68 & 20.29  \\
    mPLUG-Owl2 \citep{ye2023mplug} & 78.01 & 48.52 & 39.68 & 20.56 & 59.36 & 48.44 \\
    LLaVa-NeXT \citep{liu2024llavanext} & 68.23 & 46.74 & 41.21 & 18.43 & 39.57 & 45.57 \\
    DeepSeek-VL  \citep{lu2024deepseek} & 74.01 & 51.94 & 45.46  & 21.06 & 29.04 & 48.76\\    		
    Yi-VL \citep{young2024yi}  & 59.56 & 44.81 & 48.97 & 32.93 & 24.63 & 47.28 \\
    InternVL2-40B \citep{chen2024far} &  \textcolor{blue}{96.76} & 64.25 & 76.28  &  \textcolor{blue}{76.50} &  \textcolor{blue}{76.27} &  78.70\\    				
    \multicolumn{7}{c}{\cellcolor[HTML]{EFEFEF}Medical Special Model} \\
    MedVInT-TE \citep{zhang2023pmc}   &  62.62 & 41.03 & 40.57 & 12.17 & 45.17 & 43.83 \\	
    LLaVA-Med \citep{li2024llava}   &  48.41 & 27.96 & 23.72 & 16.10 & 21.94 & 27.82   \\ 
    Med-Flamingo \citep{moor2023med}   &  26.74 & 25.10 & 23.80 & 28.04 & 16.26 & 23.82 \\					
    RadFM \citep{wu2023towards}   &   27.45 & 21.65 & 23.75 & 16.94  & 20.05 & 23.48 \\					
    MedDr \citep{he2024meddr} & 91.37 & 51.62 & 65.56 &  73.18 & 74.52 & 68.27\\								
    HuatuoGPT-Vision-34B \citep{chen2024huatuogpt} &95.06 &  75.67 &  66.51 & 72.83 & 74.92 &  73.23 \\ 
    \multicolumn{7}{c}{\cellcolor[HTML]{FFEFEF}Our Model} \\
    \textbf{GMAI-VL (w/o our data)} &  96.40 &  \textcolor{blue}{80.97} &  \textcolor{blue}{79.14} & 70.29 &  75.66 &  \textcolor{blue}{79.96} \\
    \textbf{GMAI-VL (ours)} &  \textcolor{red}{98.64} &  \textcolor{red}{92.95} &  \textcolor{red}{88.7} & \textcolor{red}{87.21} &  \textcolor{red}{82.95} &  \textcolor{red}{88.48} \\
    \bottomrule[1pt]
    \end{tabular}%
    }
  \label{tab:omni_result}%
\end{table}%

\begin{table*}[t]
\centering
\caption{Results on the \textit{val} and \textit{test} sets of GMAI-MMBench for clinical VQA tasks. Full task names are in Table 5 of~\citep{chen2024gmai}, and additional model comparisons are in Table~\ref{tab:gmai_full} (Supplementary Materials). The best and second-best models are marked in \textcolor{red}{red} and \textcolor{blue}{blue}, respectively.}
\resizebox{0.99\linewidth}{!}{
\setlength{\tabcolsep}{3.5pt}
\begin{tabular}{l|cc|cccccccccccccccccc}
\toprule
\multirow{2}{*}{\textbf{Model Name}}  
& \multirow{2}{*}{\makecell{\textbf{Overall}\\\textbf{(val)}}}
& \multirow{2}{*}{\makecell{\textbf{Overall}\\\textbf{(test)}}}
& \multirow{2}{*}{\textbf{AR}} & \multirow{2}{*}{\textbf{BVR}} & \multirow{2}{*}{\textbf{B}} 
& \multirow{2}{*}{\textbf{CR}} & \multirow{2}{*}{\textbf{C}}  & \multirow{2}{*}{\textbf{DD}} 
& \multirow{2}{*}{\textbf{IQG}}& \multirow{2}{*}{\textbf{MR}} & \multirow{2}{*}{\textbf{M}}  
& \multirow{2}{*}{\textbf{NT}} & \multirow{2}{*}{\textbf{OR-A}}& \multirow{2}{*}{\textbf{OR-HN}} 
& \multirow{2}{*}{\textbf{OR-P}}& \multirow{2}{*}{\textbf{OR-T}}& \multirow{2}{*}{\textbf{SG}}  
& \multirow{2}{*}{\textbf{SAR}}& \multirow{2}{*}{\textbf{SIR}}& \multirow{2}{*}{\textbf{SWR}}\\
\\
\midrule
\multicolumn{21}{c}{\cellcolor[HTML]{EFEFEF}\textbf{Medical Special Model}} \\
Med-Flamingo~\citep{moor2023med}
         & 12.74 & 11.64 & 6.67  & 10.14 & 9.23  & 11.27 & 6.62  & 13.43 & 12.15
         & 6.38  & 8.00  & 18.18 & 9.26  & 18.27 & 11.00 & 11.53 & 12.16 & 5.19
         & 8.47  & 11.43 \\
LLaVA-Med~\citep{li2024llava}
         & 20.54 & 19.60 & 24.51 & 17.83 & 17.08 & 19.86 & 15.04 & 19.81 & 20.24
         & 21.51 & 13.20 & 15.15 & 20.42 & 23.73 & 17.67 & 19.65 & 21.70 & 19.81
         & 14.11 & 20.86 \\
Qilin-Med-VL-Chat~\citep{liu2023qilin}
         & 22.34 & 22.06 & 29.57 & 19.41 & 16.46 & 23.79 & 15.79 & 24.19 & 21.86
         & 16.62 & 7.20  & 13.64 & 24.00 & 14.67 & 12.67 & 15.53 & 26.13 & 24.42
         & 17.37 & 25.71 \\
RadFM~\citep{wu2023towards}
         & 22.95 & 22.93 & 27.16 & 20.63 & 13.23 & 19.14 & 20.45 & 24.51 & 23.48
         & 22.85 & 15.60 & 16.16 & 14.32 & 24.93 & 17.33 & 21.53 & 29.73 & 17.12
         & 19.59 & 31.14 \\
MedDr~\citep{he2024meddr}
         & 41.95 & 43.69 & 41.20 & 50.70 & 37.85 & 29.87 & 28.27 & 52.53 & 36.03
         & 31.45 & 29.60 & 47.47 & 33.37 & 51.33 & 32.67 & 44.47 & 35.14 & 25.19
         & 25.58 & 32.29 \\
\midrule
\multicolumn{21}{c}{\cellcolor[HTML]{EFFFFF}\textbf{Our Model}} \\
\textbf{GMAI-VL (w/o our data)}
         & \textcolor{blue}{54.99} & \textcolor{blue}{56.23} & \textcolor{blue}{51.26} 
         & \textcolor{red}{61.05} & 53.79 & 44.39 & 44.51 & \textcolor{blue}{62.60} 
         & 40.80 & \textcolor{blue}{57.42} & \textcolor{blue}{35.20} 
         & \textcolor{red}{79.50} & \textcolor{red}{61.31} & \textcolor{red}{77.81} 
         & \textcolor{red}{53.60} & \textcolor{red}{69.29} & 35.39 & \textcolor{blue}{35.77} 
         & 29.71 & \textcolor{blue}{44.86} \\

\textbf{GMAI-VL (ours)}
         & \textcolor{red}{61.74} & \textcolor{red}{62.43} & \textcolor{red}{75.26} 
         & 59.66 & \textcolor{red}{67.24} & \textcolor{red}{56.86} & \textcolor{red}{54.29} 
         & \textcolor{red}{67.14} & \textcolor{blue}{42.80} & \textcolor{red}{79.97} 
         & \textcolor{red}{41.60} & 75.00 & \textcolor{blue}{60.45} & 75.48 
         & \textcolor{blue}{53.33} & 58.12 & \textcolor{red}{42.09} & \textcolor{red}{72.31} 
         & \textcolor{red}{37.40} & \textcolor{red}{59.14} \\

\bottomrule
\end{tabular}
}
\label{tab:gmai}
\end{table*}

\begin{table*}[tp]
\centering
\caption{Evaluation on the training strategy.}
\resizebox{0.8\textwidth}{!}{
\begin{tabular}{l|c|c|c|c}
\hline
\textbf{Model} & \textbf{MMMU\_val} & \textbf{OmniMedVQA} & \textbf{GMAI\_MMBench\_val} & \textbf{GMAI\_MMBench\_test} \\ \hline
stage1 & 46.00 & 81.38 & 51.49 & 53.69 \\ \hline
stage1+2 & 46.67 & 84.64 & 55.85 & 58.28 \\ \hline
stage1+3 & 45.33 & 86.62 & 59.25 & 60.70 \\ \hline\hline
stage1+2+3 (Prop) & \textbf{51.30} & \textbf{88.48} & \textbf{61.74} & \textbf{62.43} \\ \hline
\end{tabular}}
\label{table:experiment_results}
\end{table*}

\begin{table}[tp]
\centering
\caption{Performance on the \textit{val} set for the MMMU Health \& Medicine track. 
The best performance is highlighted in \textcolor{red}{red} while the second-best performance is highlighted in \textcolor{blue}{blue}.
Note that the results are obtained from the official website.}
\setlength{\tabcolsep}{3.5pt}
\resizebox{\linewidth}{!}{
\begin{tabular}{l|ccccccc}
\toprule
\multirow{2}{*}{\textbf{Model}} & \multirow{2}{*}{\textbf{BMS}} & \multirow{2}{*}{\textbf{CM}} & \multirow{2}{*}{\textbf{DLM}} & \multirow{2}{*}{\textbf{P}} & \multirow{2}{*}{\textbf{PH}} & \multirow{2}{*}{\makecell{\textbf{MMMU} \\ \textbf{Health \& Medicine} }} \\
\\

\midrule
Med-Flamingo \citep{moor2023med} & 33.6 & 30.2 & 23.3 & 29.3 & 25.8 & 28.4 \\ 
RadFM \citep{wu2023towards} & 31.6 & 28.6 & 26.7 & 26.2 & 26.8 & 27.9 \\ 
LLaVA-Med-7B \citep{li2024llava} & 33.8 & 32.3 & 26.7 & 40.7 &\textcolor{blue}{ 43.3} & 38.6 \\ 
Qwen-VL-Chat \citep{bai2023qwen} & 32.7 & 20.6 & 19.3 & 29.6 & 33.3 & 31.7 \\ 
Yi-VL-34B \citep{young2024yi} & 48.1 & 55.6 & \textcolor{blue}{36.7} & 35.4 & 31.3 & 48.2 \\ 
LLaVA-v1.6-7B \citep{liu2023llava} & 46.4 & 43.4 & 30.0 & 29.6 & 26.7 & 33.1 \\ 
LLaVA-v1.6-13B \citep{liu2023llava} & \textcolor{red}{53.6} & 46.7 & 33.3 & 22.2 & 40.0 & 39.3 \\ 
HuatouGPT-Vision-7B \citep{chen2024huatuogpt} & \textcolor{blue}{50.0} & \textcolor{red}{63.3} & \textcolor{blue}{36.7} & \textcolor{blue}{48.1} & \textcolor{red}{53.3} & \textcolor{blue}{50.3} \\ \midrule
\hspace{0mm}\textbf{GMAI-VL(w/o our data)} & 43.3 & 56.7  & \textcolor{red}{43.3} & 46.7 & 40.0 & 46.0  \\
\hspace{0mm}\textbf{GMAI-VL(ours)} & \textcolor{blue}{50.0} & \textcolor{blue}{60.0}  & \textcolor{red}{43.3} & \textcolor{red}{50.0} & \textcolor{red}{53.3} & \textcolor{red}{51.3}  \\
\hline
\end{tabular} }
\label{tab:mmmu_result}
\vspace{-2mm}
\end{table}

\begin{figure*}[htb]
    \centering
    \includegraphics[width=0.95\textwidth]{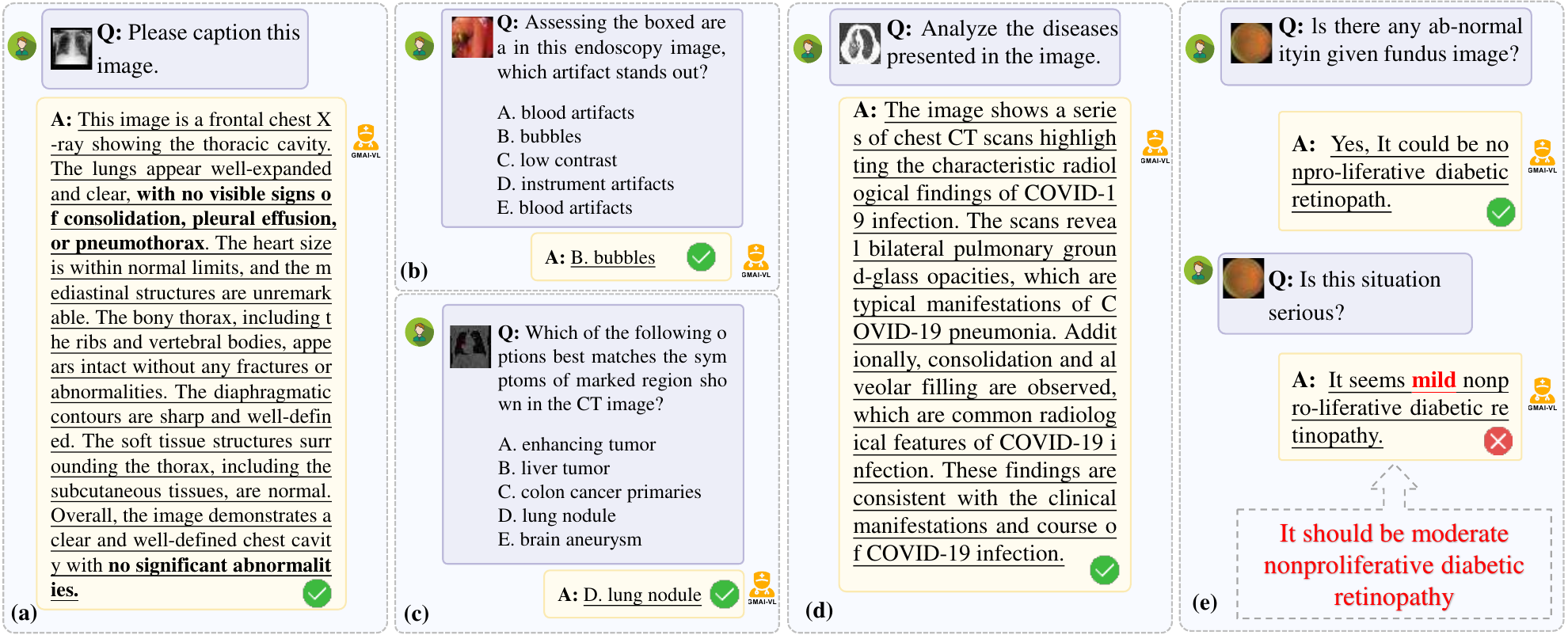}
    \vspace{-2mm}
    \caption{Example results of our GMAI-VL model. Figure (e) is a failed case.}
    \label{fig:results}
    \vspace{-2mm}
\end{figure*}



The OmniMedVQA~\citep{hu2024omnimedvqa} benchmark integrates 73 traditional medical imaging datasets, all formatted for visual question answering (VQA). 
Table~\ref{tab:omni_result} summarizes the performance of various large vision-language models (LVLMs), including GMAI-VL, across five question types: Modality Recognition, Anatomy Identification, Disease Diagnosis, Lesion Grading, and Other Biological Attributes.

From the results, we can see that (i) GMAI-VL excels in all tasks, achieving 98.64\% in Modality Recognition, 92.95\% in Anatomy Identification, and 88.71\% in Disease Diagnosis; 
(ii) it outperforms both open-source and medical-specific models, demonstrating strong abilities in identifying anatomical structures and diagnosing diseases;
(iii) in Lesion Grading, it achieves the highest score of 87.21\%, and scores 82.95\% in Other Biological Attributes, showcasing its versatility;
and
(iv) with an average accuracy of 88.48\%, GMAI-VL surpasses models like HuatuoGPT-Vision-34B and InternVL2-40B, establishing itself as a leading model in multimodal medical image understanding and setting a new benchmark for medical VQA tasks.


\subsection{Comparisons on GMAI-MMBench}

The GMAI-MMBench benchmark is a comprehensive evaluation suite for medical multimodal models, focusing on clinical visual question-answering (VQA) tasks. 
Table~\ref{tab:gmai} presents the performance of various models on the \textit{val} and \textit{test} sets across a range of clinical tasks.
Notably, (i) GMAI-VL excels in tasks such as abnormality recognition (73.78\%), biological variation recognition (63.06\%), and clinical disease diagnosis (66.67\%), demonstrating its strong ability to understand and interpret complex clinical images. 
(ii) Compared to other models, GMAI-VL consistently ranks first or second across most tasks, securing the top position in 16 out of 20 categories. 
(iii) Key tasks such as Attribute Recognition (AR) and Disease Diagnosis (DD) yield scores of 75.26\% and 67.14\%, respectively, highlighting GMAI-VL’s strength in medical scenario understanding.
(iv) Overall, GMAI-VL sets a new benchmark in various clinical VQA tasks.

\subsection{Comparisons on MMMU Health \& Medicine}

We further assess the performance of our GMAI-VL on the Health \& Medicine track of MMMU benchmark, which is a widely recognized standard for evaluating multimodal models.
The experimental results in Table~\ref{tab:mmmu_result} show the model's performance across five key categories: Basic Medical Science (\textbf{BMS}), Clinical Medicine (\textbf{CM}), Diagnostics and Laboratory Medicine (\textbf{DLM}), Pharmacy (\textbf{P}), and Public Health (\textbf{PH}).

The results show that (i) GMAI-VL performs strongly across multiple categories, achieving top scores in \textbf{DLM} (43.3\%), \textbf{P} (50.0\%), and \textbf{PH} (53.3\%), surpassing competitive models such as LLaVA-v1.6 and HuatuoGPT-Vision-7B. 
These results highlight the model's proficiency in handling complex tasks that require diagnostic reasoning, pharmaceutical knowledge, and public health expertise.
(ii) In \textbf{BMS}, GMAI-VL scores 50.0\%, achieving the best performance and demonstrating its ability to understand medical knowledge. 
(iii) In \textbf{CM}, the model scores 60.0\%, remaining competitive with other leading models. These results underscore the model's effectiveness in processing both clinical and foundational medical information.
(iv) Overall, GMAI-VL achieves an average score of 51.3\% across the Health \& Medicine track, ranking among the top models and confirming its versatility in specialized medical domains.

\subsection{Ablation Study of the GMAI-VL-5.5M Dataset}

In this section, we evaluate the effectiveness of the proposed GMAI-VL-5.5M dataset. 
We report the results of GMAI-VL (w/o our data) in Tables~\ref{tab:tradi_result}, \ref{tab:omni_result}, \ref{tab:gmai}, and \ref{tab:mmmu_result}. This model is trained using data excluding the GMAI-VL-5.5M dataset (see Sec.\ref{sec:subtraining}), highlighting that our dataset effectively enhances the performance of large vision-language models.

The results demonstrate that the GMAI-VL-5.5M dataset provides highly accurate and reliable medical knowledge, especially in recognizing and understanding multimodal medical data. This significantly improves model performance, showcasing the dataset's diversity, comprehensiveness, and its ability to complement other datasets in complex medical tasks.


\subsection{Ablation Study of Training Strategy}
Table~\ref{table:experiment_results} presents results at different training stages.
(i) The proposed stage1+2+3 (Prop) training strategy significantly enhances model performance across multiple benchmarks, outperforming other models and training strategies in all key metrics. 
(ii) By progressively incorporating stages 1, 2, and 3, the model shows consistent improvements on various datasets. 
This highlights the effectiveness of the three-stage training pipeline in improving the model's ability to handle complex multimodal medical tasks, confirming that a more comprehensive training approach yields superior results.


\subsection{Case Study}
Fig.~\ref{fig:results} shows example results of our GMAI-VL across various medical imaging and diagnostic tasks. The model accurately interprets chest X-rays, detects lesions in fundus images, and identifies COVID-19 features in chest CT scans, among others. These results highlight the model's versatility and potential in assisting clinical diagnostics. However, as illustrated in the failure case in Fig.~\ref{fig:results} (e), GMAI-VL encounters difficulties in detecting subtle differences, such as distinguishing between mild and moderate severity.



\section{Conclusion}
In this paper, we build GMAI-VL, a large vision-language model, and GMAI-VL-5.5M, a comprehensive multimodal medical dataset designed to advance general medical AI. GMAI-VL-5.5M converts hundreds of medical image analysis datasets into high-quality image-text pairs through annotation-guided data generation, enabling GMAI-VL to tackle a wide range of clinical tasks effectively. Experimental results show that GMAI-VL-5.5M significantly enhances GMAI-VL's performance, achieving state-of-the-art results across multiple key benchmark datasets. Future work will focus on expanding the dataset with more diverse and challenging medical scenarios, further improving the model's ability to generalize across different clinical environments and 
applications.
\begin{CJK}{UTF8}{gbsn}
{
\small
    \bibliographystyle{ieeenat_fullname}
    \bibliography{main}
}
\end{CJK}
\newpage
\clearpage
\def\paperID{10237} 
\def\confName{ICCV}
\def\confYear{2025}

\cleardoublepage 
\onecolumn 
\vspace*{7cm} 
\begin{center}
    {\Huge \textbf{Supplementary Material}} \\[0.5cm]
    {\LARGE \textbf{GMAI-VL \& GMAI-VL-5.5M: A Large Vision-Language Model and a Comprehensive Multimodal Dataset Towards General Medical AI}}
\end{center}
\vspace{2cm} 



\onecolumn

\vspace*{2mm}
\normalsize
\noindent
There are four parts in this supplementary material.
\ \\

\vspace{+1.6mm} \noindent 
\textbf{Part 1:} Details of GMAI-VL-5.5M

\vspace{+1.6mm} 
\noindent \textbf{Part 2:} Details of All Training Data for GMAI-VL

\vspace{+1.6mm} 
\noindent \textbf{Part 3:} Results on the Validation and Test Sets of GMAI-MMBench for Clinical VQA Tasks



\vspace{+1.6mm} 
\noindent \textbf{Part 4:} Quality Evaluation

\newpage

\section*{Part 1 - Details of GMAI-VL-5.5M}

\definecolor{blueColor}{HTML}{0054D6}

\begin{table*}[htbp]
\caption{Details of Sub-Datasets in GMAI-VL-5.5M}
\label{tab:composition}
\resizebox{1.0\textwidth}{!}{
\begin{tabular}{l|l|l|l}

\toprule
Dataset                      & Sub-Dataset Name       & Description                                                            & Size \\ \midrule
\multirow{5}{*}{GMAI-VL-5.5M} & GMAI-MM-Caption-1.7M   & A curated set of detailed medical image captions.                      & 1.7M \\ \cmidrule{2-4} 
                             & GMAI-MM-Instrunct-0.9M & A diverse set of instructions for medical image analysis.              & 0.9M \\ \cmidrule{2-4} 
                             & GMAI-MM-Percept-1.3M   & A dataset of labels for medical image classification and segmentation. & 1.3M \\ \cmidrule{2-4} 
                             & GMAI-Text-Single-1M    & A set of single-round medical dialogues on patient queries             & 1.0M \\ \cmidrule{2-4} 
                             & GMAI-Text-Multi-0.6M   & A dataset of multi-turn medical conversations on various topics.       & 0.6M \\ \bottomrule
\end{tabular}
}
\label{tab:sub_dataset}
\end{table*}

\begin{figure*}[!htbp]
    \centering
    \begin{minipage}{0.49\linewidth}
        \centering
        \includegraphics[width=0.95\linewidth]{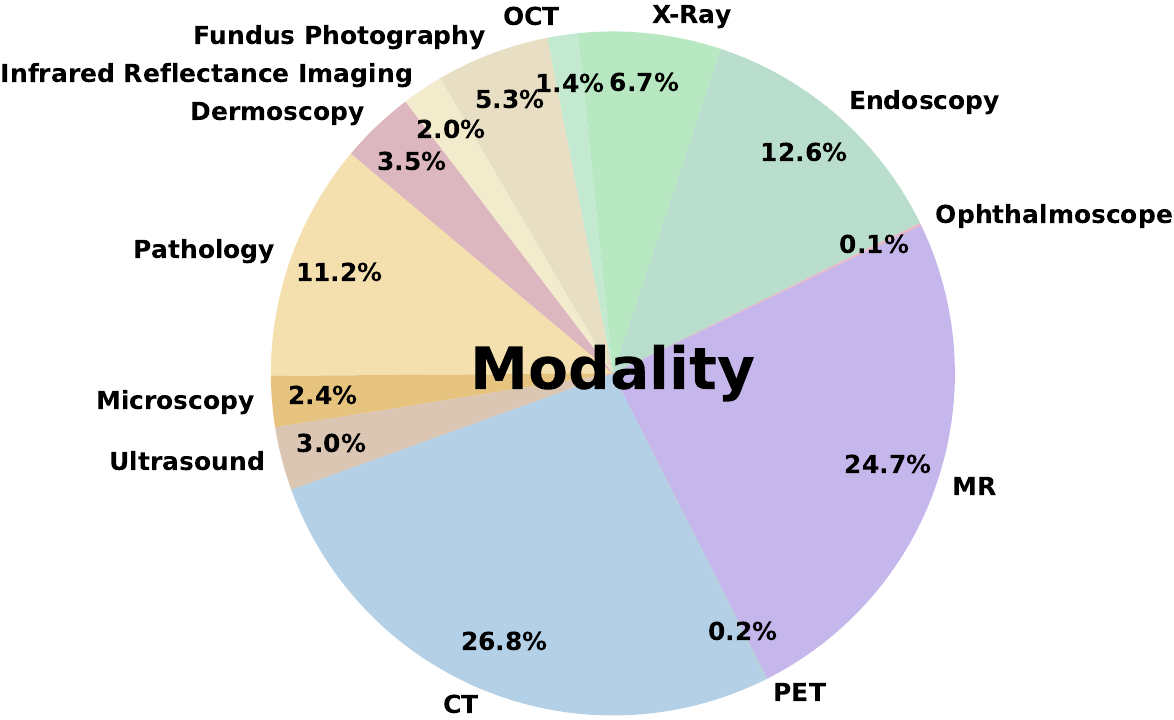}
        \subcaption*{\small(a) Modality distribution}
    \end{minipage}
    \hfill
    \begin{minipage}{0.49\linewidth}
        \centering
        \includegraphics[width=0.89\linewidth]{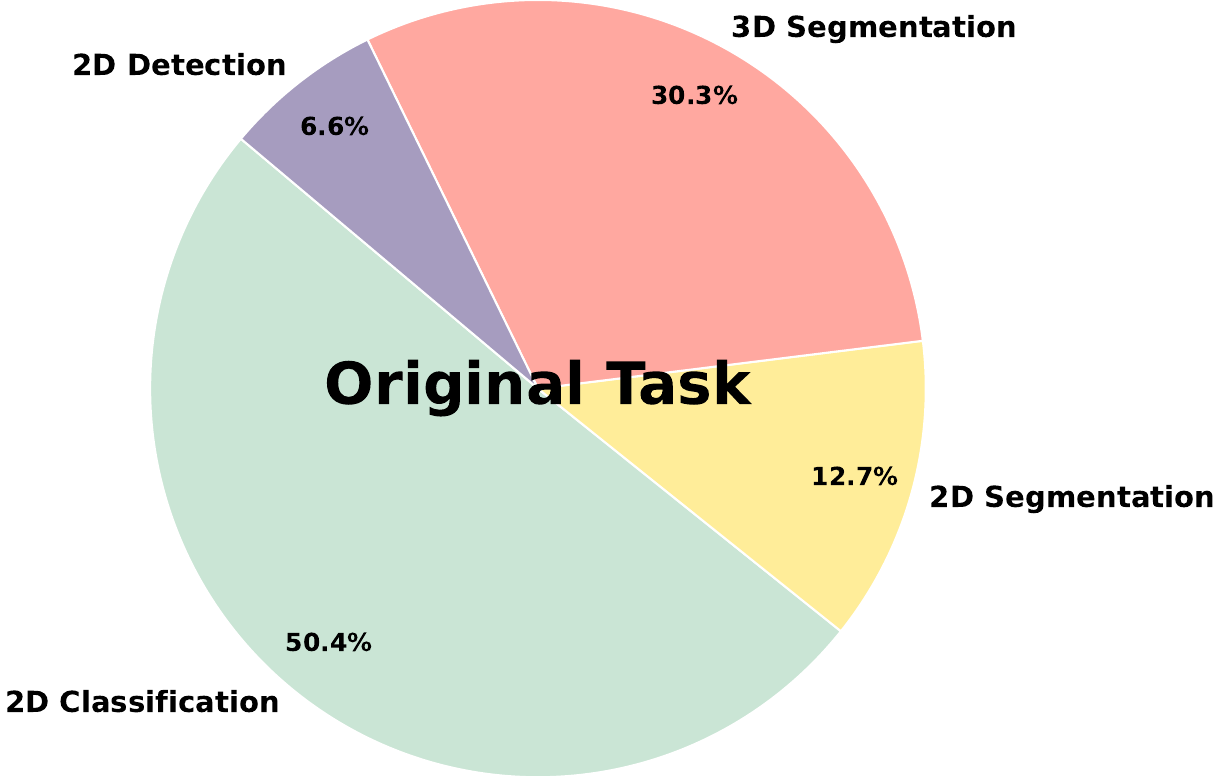}
        \subcaption*{\small(b) Original task distribution}
    \end{minipage}
    
    \vspace{0.1cm} 

    \begin{minipage}{0.49\linewidth}
        \centering
        \includegraphics[width=0.88\linewidth]{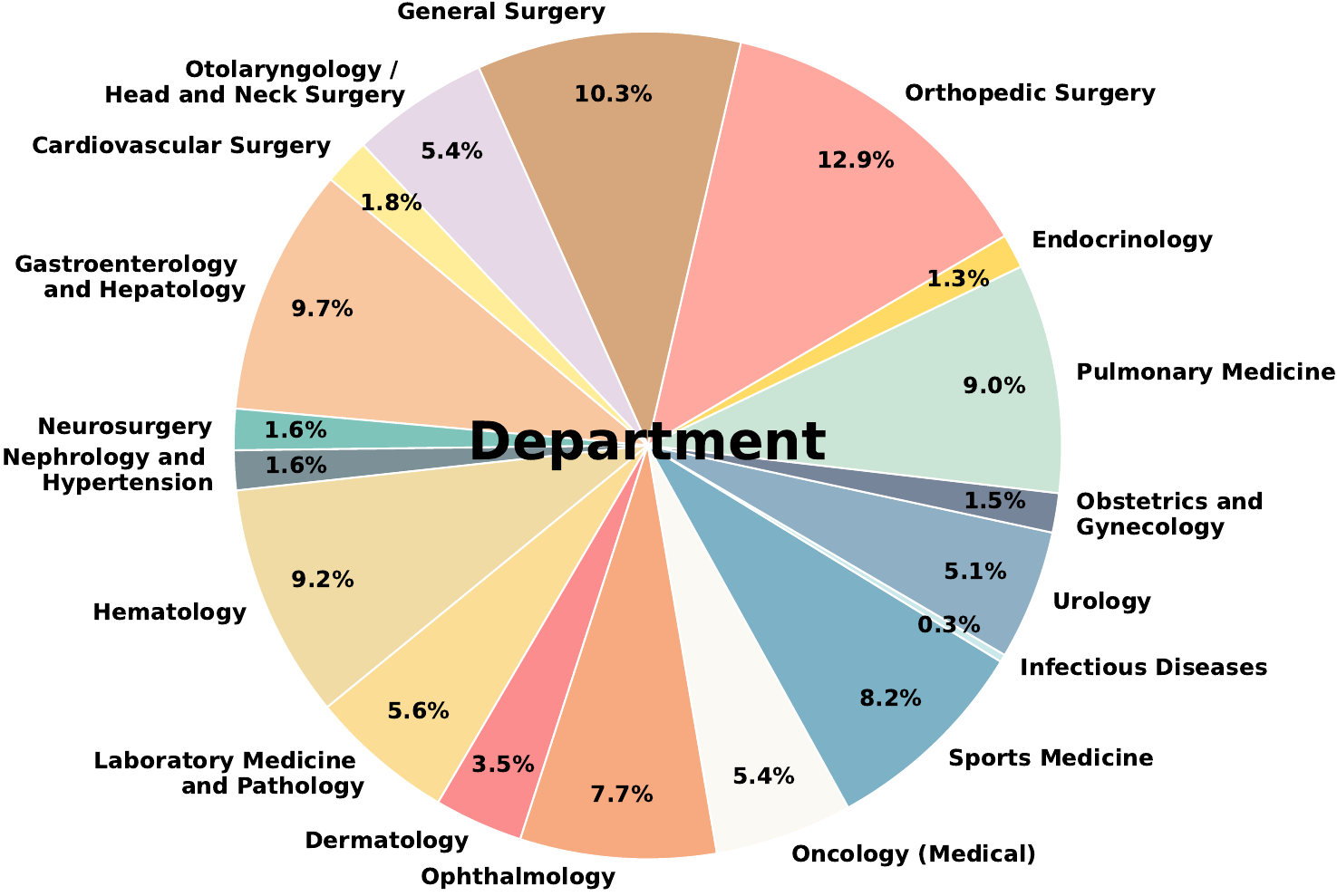}
        \subcaption*{\small(c) Department distribution}
    \end{minipage}
    \hfill
    \begin{minipage}{0.49\linewidth}
        \centering
        \includegraphics[width=\linewidth]{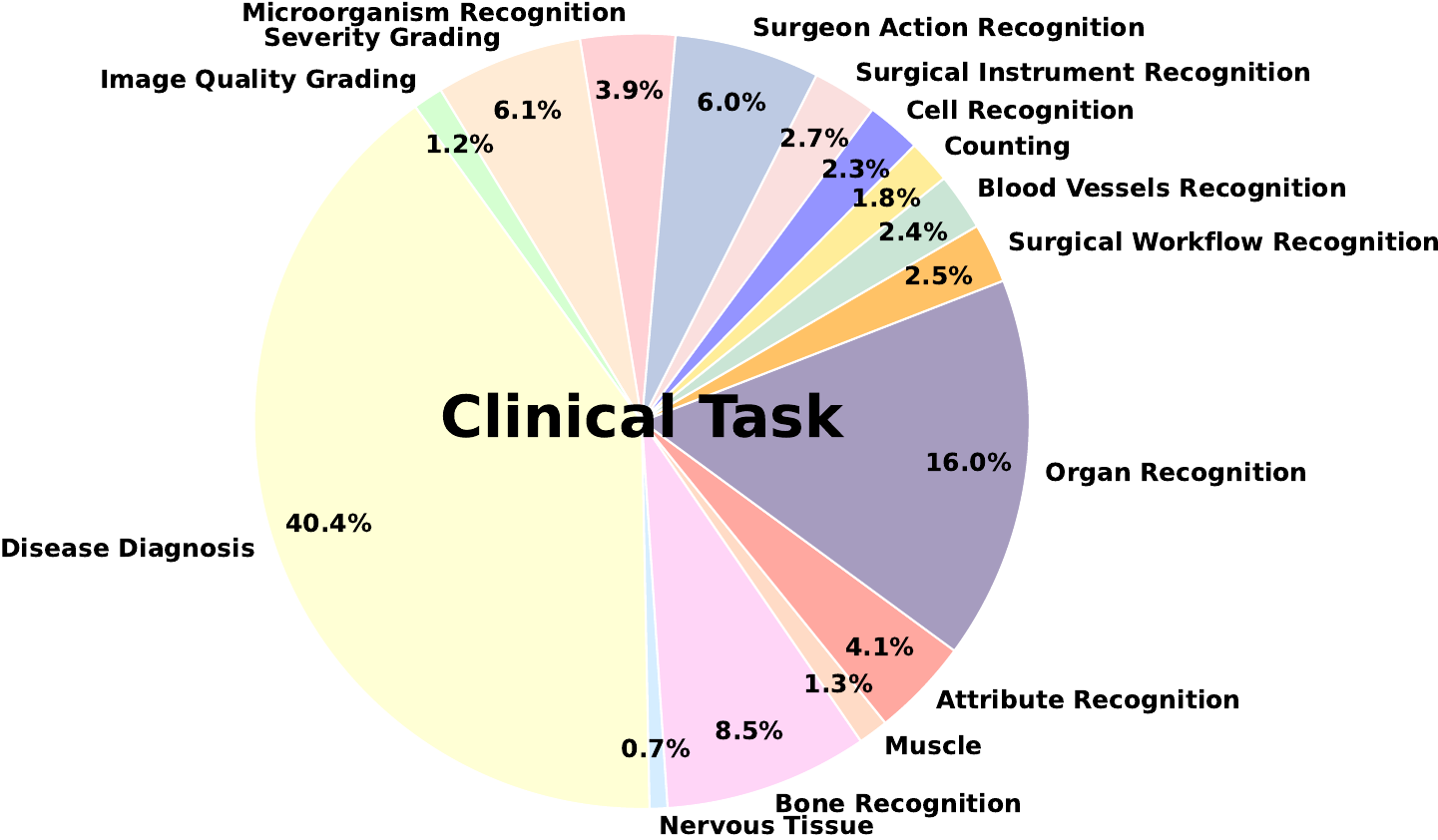}
        \subcaption*{\small(d) Clinical task distribution}
    \end{minipage}
\caption{Distribution of GMAI-VL-5.5M across various modalities, departments, and tasks.}
\label{fig:distributions}
\end{figure*}

\large

Table~\ref{tab:sub_dataset} provides information on the sub-datasets of the multimodal dataset GMAI-VL-5.5M that we have constructed. 
Based on the different data formats discussed in the paper, we have categorized the data into five distinct sub-datasets: GMAI-MM-Caption-1.7M, GMAI-MM-Instruct-0.9M, GMAI-MM-Percept-1.3M, GMAI-Text-Single-1M, and GMAI-Text-0.6M. 
Each sub-dataset corresponds to specific components, including image caption data, free instruction data, visual perception data, text-only data, and conversation data. 
Additionally, Fig.~\ref{fig:distributions} presents a comprehensive distribution of the data across different modalities, original tasks, departments, and clinical tasks within the dataset, highlighting its richness and diversity.

\newpage

\section*{Part 2 -  Details of All Training Data for GMAI-VL}

\begin{figure*}[htbp]
    \centering
    \includegraphics[width=1.0\linewidth]{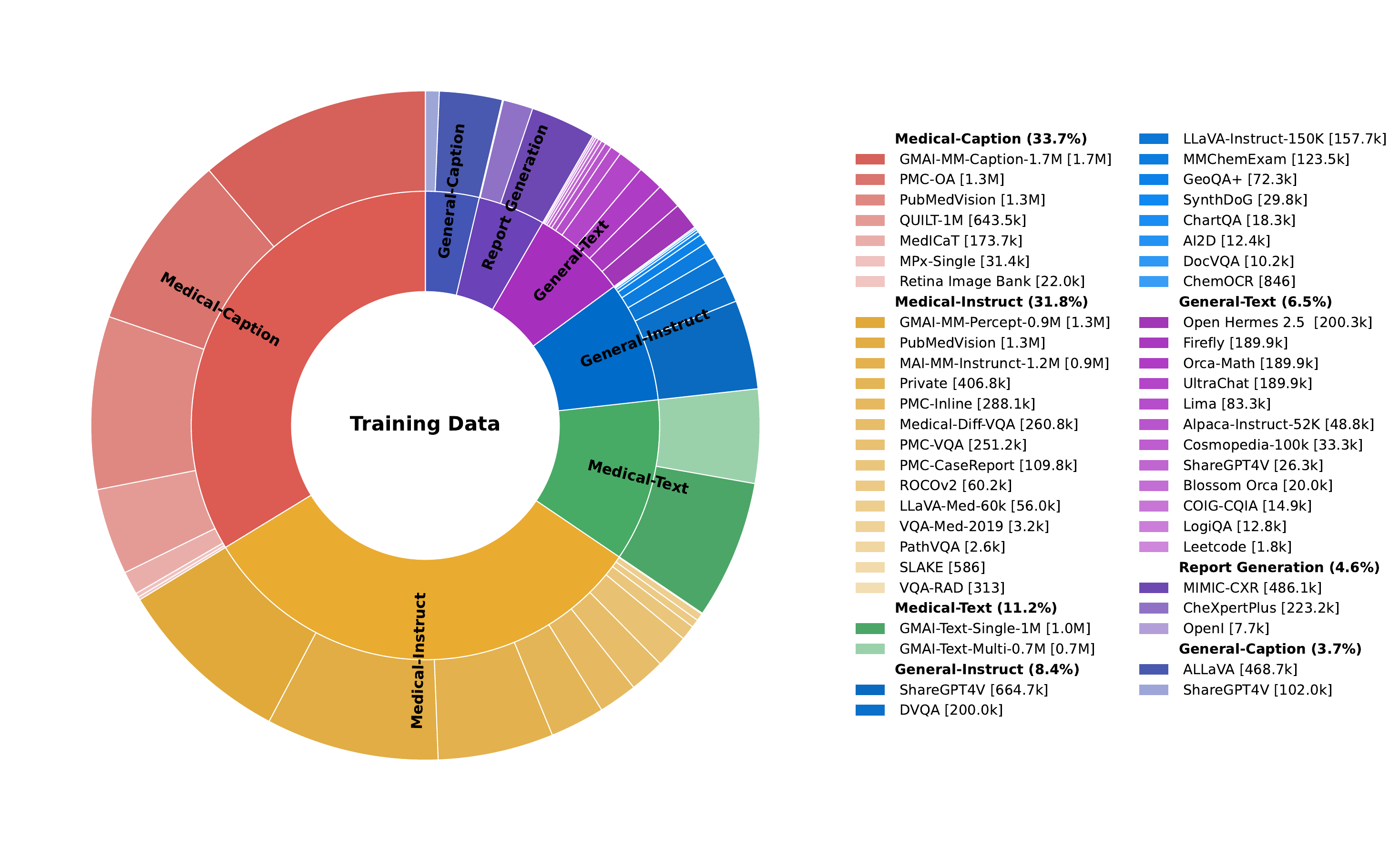}
    \caption{Distribution of the training dataset. The inner ring represents major categories, each shown in a distinct color, while the outer ring depicts the corresponding subcategories. Segment sizes are proportional to data volume, as indicated in the legend, which also provides the data volume for each subcategory.}
    \label{fig:distribution_of_our_data}
\end{figure*}

\begin{table*}[htbp]
\label{tab:training_dataset_list}
\centering
\caption{List of datasets used in our model. We employ a large collection of image-text data and instruction data for training stage.}
\label{tab:data_ratio}
    \resizebox{0.95\textwidth}{!}{
    \begin{tabular}{p{2cm}|p{6cm}|c|cc}
        \toprule
        \multirow{2}{*}{\makecell{\textbf{Dataset}\\\textbf{Category}}}  & \multirow{2}{*}{\textbf{Dataset Name}} & \multirow{2}{*}{\textbf{Size}} & \multirow{2}{*}{\textbf{ratio in stage 1\&2}} & \multirow{2}{*}{\textbf{ratio in stage 3}} \\
        & & & & \\
        \midrule
        \multirow{2}{*}{\makecell{General\\ Captioning}}  & 
        ALLaVA\citep{chen2024allava} & 468k   & \multirow{2}{*}{100.0\%} &  \multirow{2}{*}{50.0\%} \\ 
        & ShareGPT4V\citep{chen2023sharegpt4v} &102k & & \\
        
        \hline
        \multirow{7}{*}{\makecell{Medical \\Captioning}}  & GMAI-MM-Caption-1.7M   & 1.7M &\multirow{2}{*}{100.0\%} & \multirow{2}{*}{100.0\%}\\ 
         & PubMedVision\citep{chen2024huatuogpt}  & 1.3M & & \\ 
         \cline{2-5}
         & MedICaT\citep{subramanian2020medicat}  & 173k &\multirow{5}{*}{100.0\%} & \multirow{5}{*}{5.0\%}\\ 
         & MPx-Single\citep{wu2023towards}  & 31k & & \\ 
         &  PMC-OA\citep{lin2023pmc}  & 1.3M & & \\ 
         & QUILT-1M\citep{ikezogwo2024quilt}  & 643k & & \\ 
         & Retina Image Bank\citep{RetinaImageBank2024}  & 22k & & \\ 
        \hline
         
         \multirow{3}{*}{\makecell{Report \\Generation}}  &  CheXpertPlus\citep{chambon2024chexpert}  & 223k & \multirow{3}{*}{100.0\%} & \multirow{3}{*}{30.0\%} \\  
         & MIMIC-CXR\citep{johnson2019mimic}  & 486k & & \\ 
         & OpenI\citep{demner2016preparing}  & 7k & & \\
        \hline
        \multirow{8}{*}{\makecell{General\\ Instruction}}   
        & GeoQA+\citep{cao2022augmented} & 72k & \multirow{8}{*}{100.0\%} & \multirow{8}{*}{75.0\%} \\   
        &  AI2D\citep{kembhavi2016diagram} & 12k & & \\  
        &  SynthDoG\citep{kim2022ocr} & 29k & & \\   
        &  ChartQA\citep{masry2022chartqa} & 18k & & \\   
        &  MMChemExam\citep{li2024seeing} & 219k & & \\   
        &   LLaVA-Instruct-150K\citep{liu2023llava} & 157k & & \\   
        &  DVQA\citep{kafle2018dvqa} & 200k & & \\   
        &  DocVQA\citep{mathew2021docvqa} & 10k & &   \\ 
        \hline
        \multirow{13}{*}{\makecell{Medical \\ Instruction}} & GMAI-MM-Percept-1.3M & 1.3M & \multirow{4}{*}{100.0\%} & \multirow{4}{*}{100.0\%}\\ 
         &   GMAI-MM-Instruct-0.9M & 0.9M & & \\ 
         &  PubMedVision\citep{chen2024huatuogpt} & 1.28M &  &  \\ 
         &  LLaVA-Med-60k\citep{li2024llava} & 56k & & \\ 
         \cline{2-5}
         & PMC-Inline\citep{wu2023towards}  & 288k & \multirow{9}{*}{100.0\%} & \multirow{9}{*}{10.0\%}\\ 
         &  VQA-Med-2019\citep{ben2019vqa} & 3.2k & & \\ 
         &  Medical-Diff-VQA\citep{hu2023medical} & 260k & & \\ 
         &  PathVQA\citep{he2020pathvqa} & 2.6k & & \\  
         &  PMC-CaseReport\citep{wu2023towards} & 109k & & \\  
         &  PMC-VQA\citep{zhang2023pmc} & 251k & & \\ 
         &  ROCOV2\citep{ruckert2024rocov2} & 60k & & \\ 
         &  SLAKE\citep{liu2021slake} & 0.6k & & \\
         &  VQA-RAD\citep{lau2018dataset} & 0.3k & & \\ 
         
        \hline
        \multirow{12}{*}{\makecell{General Text}} & blossom\_orca\citep{blossomlm} & 20k & \multirow{12}{*}{0.0\%} & \multirow{12}{*}{100.0\%} \\  
        &  COIG-CQIA\citep{bai2024coig} & 14.8k & & \\  
        &  Cosmopedia-100k\citep{benallal2024cosmopedia} & 33k & & \\  
        &  ShareGPT4V\citep{chen2023sharegpt4v} & 26k & & \\  
        &   Orca-Math\citep{mitra2024orcamath} & 379k & & \\  
        &  Leetcode\citep{raybernard_leetcode} & 1.7k & & \\  
        &  LogiQA\citep{liu2020logiqa} & 12.7k & & \\  
        &  Lima\citep{gair_lima} & 83k & & \\  
        &  Open Hermes 2.5\citep{OpenHermes_2_5} & 200k & & \\  
        &  Firefly\citep{Firefly} & 189k & & \\  
        &   UltraChat\citep{ding2023enhancing} & 189k & & \\  
        &  Alpaca-Instruct-52K\citep{alpaca}  & 49k  & &  \\ 
        \hline
        \multirow{2}{*}{\makecell{Medical Text}}  & GMAI-Text-Single-1M & 1.0M & \multirow{2}{*}{0.0\%} & \multirow{2}{*}{100.0\%} \\  
        &  GMAI-Text-Multi-0.6M & 649k & & \\ 
        \hline
        \textbf{Overall} & - & \textbf{15.7M} &  - & - \\
        \bottomrule
    \end{tabular} }
    
\end{table*}


\large

In this part, we provide a comprehensive overview of all datasets used for training the GMAI-VL model. This includes the dataset names, categories, the amount of data used, and the proportion of training data allocated to each dataset during the three phases of model training. Fig.~\ref{fig:distribution_of_our_data} visualizes the distribution of our training data, illustrating the proportion of each category and subcategory.

Table below summarizes the datasets employed, along with their respective categories and sizes. It is important to note that for some datasets, we performed data cleaning and bilingual translation, which may result in reported dataset sizes differing from the official numbers.

\clearpage

\section*{Part 3 -  Results on the Validation and Test Sets of GMAI-MMBench for Clinical VQA Tasks}

\large 

For further comparisons, Table~\ref{tab:gmai_full} provides additional results, including a broader range of models, such as Open-Source LVLMs, Proprietary LVLMs, and Medical Special Models.

\begin{table*}[htbp]
\centering
\caption{Results on the \textit{val} and \textit{test} sets of GMAI-MMBench for clinical VQA tasks. 
The full names of the evaluated tasks can be found in Table 5 in literature\cite{chen2024gmai}. 
The best model in each category is in \textcolor{red}{red}, while the second-best is \textcolor{blue}{blue}.}
\resizebox{1.0\linewidth}{!}{
\setlength{\tabcolsep}{3.5pt}
\begin{tabular}{l|cc|cccccccccccccccccc}
\toprule
\multirow{2}{*}{\textbf{Model Name}}  
& \multirow{2}{*}{\makecell{\textbf{Overall}\\\textbf{(val)}}}
& \multirow{2}{*}{\makecell{\textbf{Overall}\\\textbf{(test)}}}
& \multirow{2}{*}{\textbf{AR}} & \multirow{2}{*}{\textbf{BVR}} & \multirow{2}{*}{\textbf{B}} 
& \multirow{2}{*}{\textbf{CR}} & \multirow{2}{*}{\textbf{C}}  & \multirow{2}{*}{\textbf{DD}} 
& \multirow{2}{*}{\textbf{IQG}}& \multirow{2}{*}{\textbf{MR}} & \multirow{2}{*}{\textbf{M}}  
& \multirow{2}{*}{\textbf{NT}} & \multirow{2}{*}{\textbf{OR-A}}& \multirow{2}{*}{\textbf{OR-HN}} 
& \multirow{2}{*}{\textbf{OR-P}}& \multirow{2}{*}{\textbf{OR-T}}& \multirow{2}{*}{\textbf{SG}}  
& \multirow{2}{*}{\textbf{SAR}}& \multirow{2}{*}{\textbf{SIR}}& \multirow{2}{*}{\textbf{SWR}}\\
\\
\midrule
\multicolumn{21}{c}{\cellcolor[HTML]{EFEFFF}\textbf{Random Guess}} \\
Random   & 25.70 & 25.94 & 38.20 & 22.73 & 22.92 & 22.72 & 24.06 & 26.66 & 27.13 
         & 27.00 & 20.00 & 24.75 & 21.37 & 22.93 & 22.33 & 21.18 & 32.43 & 24.23 
         & 21.39 & 23.71 \\
\midrule
\multicolumn{21}{c}{\cellcolor[HTML]{FFF3E4}\textbf{Open-Source LVLMs}} \\
Flamingo v2\cite{awadalla2023openflamingo}
         & 25.58 & 26.34 & 37.74 & 21.50 & 20.62 & 22.00 & 22.41 & 27.29 & 25.91
         & 27.45 & 18.00 & 28.79 & 25.16 & 22.13 & 22.00 & 22.00 & 34.61 & 22.88
         & 20.44 & 27.43 \\
VisualGLM-6B\cite{ding2021cogview}
         & 29.58 & 30.45 & 40.16 & 33.92 & 24.92 & 25.22 & 24.21 & 32.99 & 29.96
         & 29.53 & 21.20 & 37.88 & 30.32 & 24.80 & 13.33 & 29.88 & 33.11 & 19.62
         & 19.16 & 37.43 \\
InstructBLIP-7B\cite{dai2024instructblip}
         & 31.80 & 30.95 & 42.12 & 26.92 & 24.92 & 28.09 & 21.65 & 34.58 & 31.58
         & 29.23 & 22.40 & 30.30 & 28.95 & 27.47 & 23.00 & 24.82 & 32.88 & 19.81
         & 21.64 & 26.57 \\
Qwen-VL\cite{bai2023qwen}
         & 34.80 & 36.05 & 37.05 & 37.24 & 35.85 & 28.98 & 24.81 & 43.60 & 24.70
         & 30.12 & 19.20 & 44.44 & 29.68 & 31.87 & 25.00 & 31.18 & 30.26 & 21.54
         & 20.10 & 26.86 \\
Yi-VL-6B\cite{young2024yi}
         & 34.82 & 34.31 & 41.66 & 39.16 & 26.62 & 30.23 & 31.88 & 38.01 & 26.72
         & 24.93 & 25.20 & 37.37 & 29.58 & 31.20 & 32.33 & 30.59 & 36.71 & 24.81
         & 23.18 & 31.43 \\
ShareGPT4V-7B\cite{chen2023sharegpt4v}
         & 36.71 & 36.70 & 43.96 & 37.59 & 21.54 & 37.57 & 18.80 & 43.26 & 32.39
         & 27.30 & 22.80 & 43.43 & 29.47 & 37.33 & 22.00 & 31.76 & 34.98 & 24.42
         & 25.06 & 30.00 \\
LLAVA-V1.5-7B\cite{liu2023llava}
         & 38.23 & 37.96 & 45.45 & 34.27 & 30.92 & 41.32 & 21.65 & 44.68 & 34.01
         & 27.74 & 23.60 & 43.43 & 28.00 & 42.13 & 29.00 & 35.06 & 33.41 & 22.12
         & 23.61 & 29.14 \\
XComposer2\cite{internlmxcomposer2}
         & 38.68 & 39.20 & 41.89 & 37.59 & 33.69 & 40.79 & 22.26 & 45.87 & 36.44
         & 32.94 & 27.20 & 58.59 & 26.11 & 36.40 & 43.67 & 37.29 & 32.06 & 23.46
         & 27.80 & 32.86 \\
LLAVA-InternLM-7b\cite{2023xtuner}
         & 38.71 & 39.11 & 36.36 & 36.54 & 32.62 & 38.10 & 30.68 & 46.53 & 34.82
         & 28.19 & 25.20 & 48.99 & 28.11 & 40.53 & 33.33 & 36.00 & 34.08 & 26.73
         & 24.12 & 29.71 \\
InternVL-Chat-V1.5\cite{chen2024far}
         & 38.86 & 39.73 & 43.84 & 44.58 & 34.00 & 33.99 & 31.28 & 45.59 & 33.20
         & 38.28 & 32.40 & 42.42 & 31.89 & 42.80 & 27.00 & 36.82 & 34.76 & 23.27
         & 24.72 & 32.57 \\
InternVL-Chat-V1.2\cite{chen2023internvl}
         & 39.52 & 40.01 & 41.66 & 44.06 & 27.38 & 38.46 & 34.29 & 46.99 & 33.60
         & 34.42 & 21.20 & 47.98 & 30.63 & 42.80 & 27.67 & 35.88 & 35.59 & \textcolor{blue}{23.85}
         & 24.98 & 28.00 \\
LLAVA-InternLM2-7b\cite{2023xtuner}
         & 40.07 & 40.45 & 39.82 & 37.94 & 30.62 & 35.24 & 29.77 & 48.97 & 34.01
         & 25.96 & 20.80 & 53.03 & 30.95 & 42.67 & 32.00 & 39.88 & 32.43 & 21.73
         & 24.38 & 38.00 \\
DeepSeek-VL-7B\cite{lu2024deepseek}
         & 41.73 & 43.43 & 38.43 & 47.03 & 42.31 & 37.03 & 26.47 & 51.11 & 33.20
         & 31.16 & 26.00 & 44.95 & 36.00 & 58.13 & 36.33 & 47.29 & 34.91 & 18.08
         & 25.49 & \textcolor{blue}{39.43} \\
MiniCPM-V2\cite{xu2024llava-uhd}
         & 41.79 & 42.54 & 40.74 & 43.01 & 36.46 & 37.57 & 27.82 & 51.08 & 28.74
         & 29.08 & 26.80 & 47.47 & 37.05 & 46.40 & 25.33 & 46.59 & 35.89 & 22.31
         & 23.44 & 31.71 \\
\midrule
\multicolumn{21}{c}{\cellcolor[HTML]{FFF0F0}\textbf{Proprietary LVLMs}} \\
Claude3-Opus\cite{anthropic2024claude}
         & 32.37 & 32.44 & 1.61  & 39.51 & 34.31 & 31.66 & 12.63 & 39.26 & 28.74
         & 30.86 & 22.40 & 37.37 & 25.79 & 41.07 & 29.33 & 33.18 & 31.31 & 21.35
         & 23.87 & 4.00 \\
Qwen-VL-Max\cite{bai2023qwen}
         & 41.34 & 42.16 & 32.68 & 44.58 & 31.38 & 40.79 & 10.68 & 50.53 & 32.79
         & 44.36 & 29.20 & 51.52 & 41.37 & 58.00 & 30.67 & 41.65 & 26.95 & 25.00
         & 24.64 & 39.14 \\
GPT-4V\cite{achiam2023gpt}
         & 42.50 & 44.08 & 29.92 & 48.95 & 44.00 & 37.39 & 12.93 & 52.88 & 32.79
         & 44.21 & \textcolor{blue}{32.80} & 63.64 & 39.89 & 54.13 & 37.00 & 50.59
         & 27.55 & 23.08 & 25.75 & 37.43 \\
Gemini 1.0\cite{team2023gemini}
         & 44.38 & 44.93 & \textcolor{blue}{42.12} & 45.10 & 46.46 & 37.57 & 20.45
         & 53.29 & 35.22 & 36.94 & 25.20 & 51.01 & 34.74 & 59.60 & 34.00 & 50.00
         & \textcolor{blue}{36.64} & 23.65 & 23.87 & 35.43 \\
Gemini 1.5\cite{reid2024gemini}
         & \textcolor{blue}{47.42} & \textcolor{blue}{48.36} & 43.50 & \textcolor{blue}{56.12}
         & \textcolor{blue}{51.23} & \textcolor{blue}{47.58} & 2.26  & \textcolor{blue}{55.33}
         & \textcolor{blue}{38.87} & \textcolor{blue}{48.07} & 30.00 & \textcolor{blue}{76.26}
         & \textcolor{blue}{51.05} & \textcolor{blue}{75.87} & 46.33 & 62.24 & 20.57 & 27.69
         & \textcolor{blue}{30.54} & 40.57 \\
GPT-4o\cite{achiam2023gpt}
         & 53.53 & 53.96 & 38.32 & \textcolor{blue}{61.01} & \textcolor{blue}{57.08} 
         & \textcolor{blue}{49.02} & \textcolor{blue}{46.62} & 61.45 
         & \textcolor{red}{46.56} & 56.38 & 34.00 & 75.25 & 53.79 & 69.47 & 48.67
         & \textcolor{blue}{65.88} & 33.93 & 22.88 & 29.51 & 39.43 \\
\midrule
\multicolumn{21}{c}{\cellcolor[HTML]{EFEFEF}\textbf{Medical Special Model}} \\
Med-Flamingo\cite{moor2023med}
         & 12.74 & 11.64 & 6.67  & 10.14 & 9.23  & 11.27 & 6.62  & 13.43 & 12.15
         & 6.38  & 8.00  & 18.18 & 9.26  & 18.27 & 11.00 & 11.53 & 12.16 & 5.19
         & 8.47  & 11.43 \\
LLaVA-Med\cite{li2024llava}
         & 20.54 & 19.60 & 24.51 & 17.83 & 17.08 & 19.86 & 15.04 & 19.81 & 20.24
         & 21.51 & 13.20 & 15.15 & 20.42 & 23.73 & 17.67 & 19.65 & 21.70 & 19.81
         & 14.11 & 20.86 \\
Qilin-Med-VL-Chat\cite{liu2023qilin}
         & 22.34 & 22.06 & 29.57 & 19.41 & 16.46 & 23.79 & 15.79 & 24.19 & 21.86
         & 16.62 & 7.20  & 13.64 & 24.00 & 14.67 & 12.67 & 15.53 & 26.13 & 24.42
         & 17.37 & 25.71 \\
RadFM\cite{wu2023towards}
         & 22.95 & 22.93 & 27.16 & 20.63 & 13.23 & 19.14 & 20.45 & 24.51 & 23.48
         & 22.85 & 15.60 & 16.16 & 14.32 & 24.93 & 17.33 & 21.53 & 29.73 & 17.12
         & 19.59 & 31.14 \\
MedDr\cite{he2024meddr}
         & 41.95 & 43.69 & 41.20 & 50.70 & 37.85 & 29.87 & 28.27 & 52.53 & 36.03
         & 31.45 & 29.60 & 47.47 & 33.37 & 51.33 & 32.67 & 44.47 & 35.14 & 25.19
         & 25.58 & 32.29 \\
\midrule
\multicolumn{21}{c}{\cellcolor[HTML]{EFFFFF}\textbf{Our Model}} \\
\textbf{GMAI-VL(w/o our data)}
         & \textcolor{blue}{54.99} & \textcolor{blue}{56.23} & \textcolor{blue}{51.26} 
         & \textcolor{red}{61.05} & 53.79 & 44.39 & 44.51 & \textcolor{blue}{62.60} 
         & 40.80 & \textcolor{blue}{57.42} & \textcolor{blue}{35.20} 
         & \textcolor{red}{79.50} & \textcolor{red}{61.31} & \textcolor{red}{77.81} 
         & \textcolor{red}{53.60} & \textcolor{red}{69.29} & 35.39 & \textcolor{blue}{35.77} 
         & 29.71 & \textcolor{blue}{44.86} \\

\textbf{GMAI-VL(ours)}
         & \textcolor{red}{61.74} & \textcolor{red}{62.43} & \textcolor{red}{75.26} 
         & 59.66 & \textcolor{red}{67.24} & \textcolor{red}{56.86} & \textcolor{red}{54.29} 
         & \textcolor{red}{67.14} & \textcolor{blue}{42.80} & \textcolor{red}{79.97} 
         & \textcolor{red}{41.60} & 75.00 & \textcolor{blue}{60.45} & 75.48 
         & \textcolor{blue}{53.33} & 58.12 & \textcolor{red}{42.09} & \textcolor{red}{72.31} 
         & \textcolor{red}{37.40} & \textcolor{red}{59.14} \\

\bottomrule
\end{tabular}
}
\label{tab:gmai_full}
\end{table*}

\clearpage

\section*{Part 4 - Quality Evaluation}
\label{sec:check}

\begin{table}[h]
    \centering
    \caption{Prompt for Evaluating the Quality of Medical Image-Text Dialogues}
    \label{tab:quality}  
        \begin{tabular}{|p{\textwidth}|}
        \hline
        You are a medical imaging expert tasked with scoring each medical text-image dialogue according to the following criteria. The score range is from 1 to 5, where a lower score indicates higher dialogue quality. Each dialogue should be evaluated based on the following criteria, considering language clarity, completeness, medical accuracy, and relevance to medical imaging. Below is an example of the dialogue content and related information: \\
        \textbf{- Modality:} \textless image modality\textgreater (e.g., CT, MRI, X-ray, etc.) \\
        \textbf{- Department:} \textless department\textgreater (e.g., Ophthalmology, Pulmonary etc.) \\
        \textbf{- Label:} \textless label\textgreater (e.g., The pre-defined label belong to image) \\
        \textbf{- Dialogue Content:} \\
        \quad - \textbf{Question:} \textless question\textgreater Question content\textless/question\textgreater \\
        \quad - \textbf{Answer:} \textless answer\textgreater Answer content\textless /answer\textgreater \\

        Please score each dialogue based on the following dimensions: \\
        \textbf{1. Medical Information Accuracy:} \\
        \quad - Ensure that medical terminology, diagnostic processes, and treatment plans align with current medical standards and practices. \\
        \quad - Assess whether there are any medical errors or misleading information that could impact diagnostic or treatment decisions. \\
        \quad - \textit{Example: If an imaging description or medical terminology is incorrect, how does this affect the overall evaluation?} \\

        \textbf{2. Language Clarity and Fluency:} \\
        \quad - Evaluate whether the dialogue is concise, clear, and easy to understand. \\
        \quad - Assess whether the language is fluent and natural, and if there are any overly complex sentence structures that might hinder information transmission. \\
        \quad - \textit{Example: Are there any unnecessary words or overly complex structures that slow down the understanding of the content?} \\
        \textbf{3. Dialogue Completeness:} \\
        \quad - Evaluate whether the dialogue includes complete medical information, covering the patient's history, imaging results, diagnostic information, and any necessary details. \\
        \quad - Assess whether any essential information is missing, omitted, or incomplete. \\
        \quad - \textit{Example: If imaging results are not mentioned or the information is incomplete, does this affect the overall evaluation?} \\
        \textbf{4. Medical Imaging Relevance:} \\
        \quad - Evaluate whether the analysis of medical imaging information is thorough and accurate. \\
        \quad - Ensure that the imaging data is appropriately interpreted, the descriptions are clear and accurate, and they align with the medical diagnosis. \\
        \quad - Assess whether the imaging data effectively supports or supplements other medical information. \\
        \quad - \textit{Example: If the description of the imaging data is inaccurate or incomplete, does it compromise the reliability of the diagnostic outcome?} \\
        \textbf{Scoring Criteria:} \\
        - 1: The dialogue is concise, accurate, and clear, fully conforming to medical standards. The imaging interpretation is appropriate, the information is complete and accurate, and it fully supports the medical diagnosis. \\
        - 2: The dialogue is generally accurate, the language is clear, and the content is complete. There may be slight imperfections in expression or minor inaccuracies in imaging interpretation, but overall quality is high. \\
        - 3: The dialogue is accurate and complete, but there may be slight room for improvement in certain areas. For example, the imaging interpretation could be clearer, or language expression could be slightly more refined. Overall quality is good and suitable for training data. \\
        - 4: The dialogue contains some notable medical errors or unclear language that affects accuracy. The imaging interpretation is problematic, or it does not integrate well with other clinical information, leading to incomplete content and a reduction in quality. \\
        - 5: The dialogue contains errors or is unclear, potentially misleading medical decisions. There is a lack of effective imaging interpretation, the information is incomplete, and the language is unclear, which compromises the medical diagnosis. The overall quality is poor. \\
        \\

        \textbf{Output Format:} \\
        Please output the results in JSON format as follows: \\
        \{"score": "Quality score", "comments": "the comments for the quality criteria",\}
        \\
        \hline
    \end{tabular}
\end{table}

To ensure the quality of the generated image-text pairs, we propose a systematic evaluation process based on a 5-level scoring system. The definition of each scoring level is outlined in the prompt (Table~\ref{tab:quality}). 
The primary objective of this process is to assess the quality of image-text pairs from a collection of 219 generated datasets and 42 publicly available open-source datasets. 
To ensure the representativeness and scientific rigor of the evaluation, we adopt a \textbf{sampling survey} methodology.
Specifically, we randomly select a subset of image-text pairs from each dataset for quality assessment, from which we infer the overall quality of the entire dataset. This sampling survey strategy enables us to efficiently and scientifically evaluate the quality of the image-text pairs, maximizing the reliability and accuracy of the results while significantly reducing the manual review workload.

The evaluation process is conducted in multiple stages:

\begin{enumerate}
    \item \textbf{Initial Quality Assessment:} From each dataset, 30 random image-text pairs are selected for evaluation. The first step involves an initial quality screening using a multimodal large language model (such as GPT-4o) to automatically assess the quality of the pairs. The model utilizes a predefined prompt, as illustrated in Table~\ref{tab:quality}, to assign preliminary quality scores based on the defined criteria (accuracy of medical information, clarity and fluency of language, completeness of the dialogue, and relevance to medical imaging).
    
    \item \textbf{Expert Review:} Following the model's initial assessment, the 30 image-text pairs are then reviewed by five medical imaging experts. 
    These experts validate the preliminary scores and provide their independent judgments on the quality of the pairs, ensuring that the evaluation aligns with current medical standards and practices.
    
    \item \textbf{Final Scoring:} After expert review, the final score for each dataset is determined based on the consensus of the expert evaluations. If the average score of the 30 image-text pairs from a dataset is less than or equal to 3 (on a scale of 1 to 5), the dataset is classified as having satisfactory quality.
\end{enumerate}

\noindent\textbf{Conclusion}: After rigorous evaluation, \textbf{95\% (247/261)} of the datasets met the predefined quality standards and were deemed acceptable. Our analysis showed that most datasets generated by GPT using the given labels exhibited high quality, with only a few requiring prompt optimization and additional information to meet the desired standards. For low-quality datasets, we addressed the issues by refining the prompt design, improving label specificity, and incorporating additional contextual information to ensure the generated quality met the requirements.
In contrast, 14 open-release multimodal medical datasets, such as, PMC-OA\cite{lin2023pmc}, quilt-1m\cite{ikezogwo2023quilt}, often extracted from research papers or manually converted, exhibited various issues such as incomplete dialogues, misalignment between images and text, overly brief responses, inclusion of non-medical images, and poor image clarity. Due to these limitations, only a portion of the open-source datasets met the required quality standards.

This systematic approach ensures that only high-quality image-text pairs are included for further use in medical imaging research, with clear documentation of the evaluation process and criteria.


\clearpage
\begin{CJK}{UTF8}{gbsn}
\end{CJK}
\end{document}